\documentclass[10pt,twocolumn,letterpaper]{article}

\usepackage{wacv}
\usepackage{epsfig}

\usepackage{balance}

\usepackage{dblfloatfix}

\usepackage{grffile}

\usepackage{wrapfig}

\usepackage{placeins}

\usepackage[inline]{enumitem}

\usepackage{xargs}

\usepackage{xspace}

\usepackage{graphicx}
\usepackage{subcaption}

\usepackage{times}
\usepackage{amsmath,amssymb} 

\usepackage{color}
\usepackage{bm}
\usepackage{mathtools}

\usepackage{tabu}

\usepackage[bookmarks=false]{hyperref}
\usepackage[hyphenbreaks]{breakurl}

\usepackage[hyperref=true,backend=biber,style=ieee,citestyle=numeric-comp]{biblatex}
\usepackage[utf8]{inputenc}

\usepackage{textcomp}

\usepackage{wrapfig}

\usepackage{floatrow}
\newfloatcommand{capbtabbox}{table}[][\FBwidth]

\newcommandx*{\skel}[1]{\ensuremath{#1}\xspace}
\newcommandx*{\ptThreeD}[2][1=i, 2=X, usedefault=@]{\ensuremath{\bm{#2}_{#1}}\xspace}
\newcommandx*{\makebold}[1]{\ensuremath{\bm{#1}}\xspace}
\newcommandx*{\plane}[1][1=$\pi$]{\makebold{#1}}
\newcommandx*{\normal}[1][1=n]{\ensuremath{\hat{\makebold{#1}}}\xspace}
\newcommandx*{\bmu}{\makebold{\mu}}
\newcommandx*{\symThreeD}{\skel{s_\text{3D}}}
\newcommandx*{\tThreeD}{\skel{t_\text{3D}}}
\newcommandx*{\symTwoD}{\skel{s_\text{2D}}}
\newcommandx*{\tTwoD}{\skel{t_\text{2D}}}
\newcommandx*{\param}{\skel{\theta}}
\newcommandx*{\Param}{\skel{\Theta}}
\newcommandx*{\Domain}{\skel{\mathcal{X}}}
\newcommandx*{\Image}{\skel{\mathcal{Y}}}

\newcommand{\figref}[1]{Fig.~\ref{#1}}

\hypersetup{
  pagebackref=false,
  colorlinks=true,
  bookmarks=false
}

\begin{document}

\wacvfinalcopy 

\def\wacvPaperID{***} 
\def\httilde{\mbox{\tt\raisebox{-.5ex}{\symbol{126}}}}

\ifwacvfinal\pagestyle{empty}\fi
\setcounter{page}{1}

\title{A Self-Supervised Bootstrap Method for Single-Image 3D Face Reconstruction}

\author{Yifan Xing \hspace{2cm} Rahul Tewari \hspace{2cm} Paulo R.\ S.\ Mendon\c{c}a \\\\
  Amazon Web Services \\
  {\tt\small yifax@amazon.com}
}

\maketitle

\ifwacvfinal\thispagestyle{empty}\fi

\begin{abstract}
  State-of-the-art methods for 3D reconstruction of faces from a single image require 2D-3D pairs of ground-truth data for supervision.
  Such data is costly to acquire, and most datasets available in the literature are restricted to pairs for which the input 2D images depict faces in a near fronto-parallel pose. Therefore, many data-driven methods for single-image 3D facial reconstruction perform poorly on profile and near-profile faces. We propose a method to improve the performance of single-image 3D facial reconstruction networks by utilizing the network to synthesize its own training data for fine-tuning, comprising:
  \begin{enumerate*}[label=(\roman*)]
    \item single-image 3D reconstruction of faces in near-frontal images without ground-truth 3D shape;
    \item application of a rigid-body transformation to the reconstructed face model;
    \item rendering of the face model from new viewpoints; and
    \item use of the rendered image and corresponding 3D reconstruction as additional data for supervised fine-tuning.
  \end{enumerate*}
  The new 2D-3D pairs thus produced have the same high-quality observed for near fronto-parallel reconstructions, thereby nudging the network towards more uniform performance as a function of the viewing angle of input faces. Application of the proposed technique to the fine-tuning of a state-of-the-art single-image 3D-reconstruction network for faces demonstrates the usefulness of the method, with particularly significant gains for profile or near-profile views. 
\end{abstract}
\begin{figure*}
	\includegraphics[width=\textwidth]{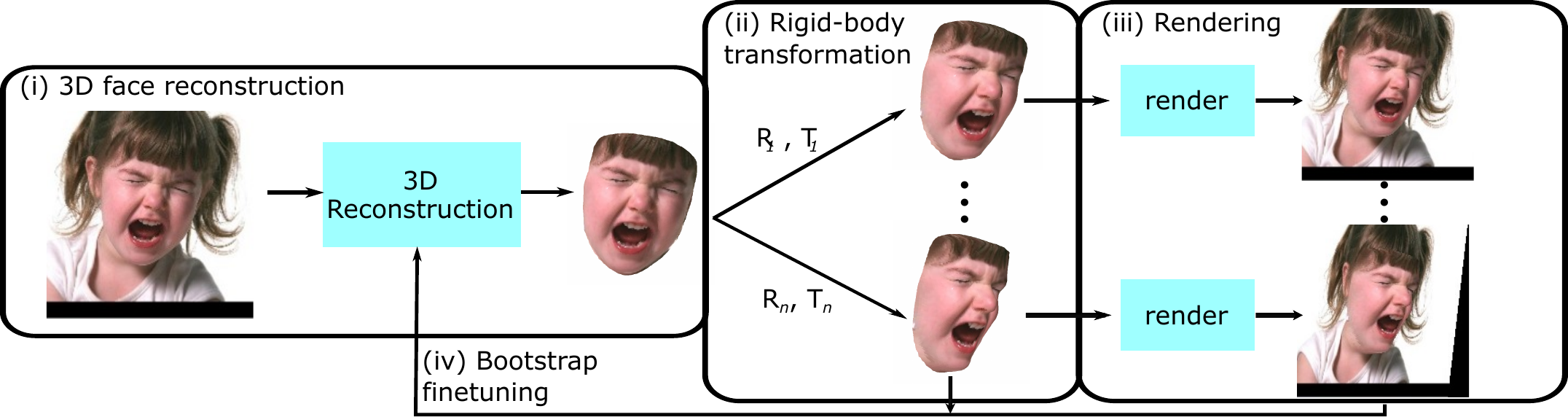}
	\caption{\label{fig:method_overview} Overview of the proposed self-supervised fine-tuning method: 1. Starting from any near-frontal image of a face, a 3D model is reconstructed using any given reconstruction network. 2. A series of rigid-body transformations are applied to the reconstructed model. 3. A set of new images are produced by rendering novel views of the model. 4. The newly rendered images and the transformed models serve as additional supervising data for fine-tuning the original reconstruction method.}
\end{figure*}
\section{Introduction\label{sec:introduction}}

Progress in deep-learning methods has enabled the solution of ill-posed problems such as single-image 3D reconstruction.
In particular, promising results have been demonstrated on single-image reconstruction of faces \cite{DBLP:journals/corr/abs-1802-00542, JacksonEtAl:FaceRecon:ICCV:2017, DBLP:journals/corr/ZhuLLSL15, DBLP:journals/corr/KimZTTRT17, DBLP:journals/corr/TranHMM16, DBLP:journals/corr/abs-1712-02859, Jourabloo2016LargePoseFA}.
The majority of these methods require 2D-3D ground-truth pairs for supervision during training.
However, ground-truth 3D shapes for faces are usually costly to acquire and the collected data often displays an imbalanced long-tail pose distribution, typically oversampling near-frontal views, 
leading to poor performance at non-frontal views.

We propose a self-supervised bootstrap method that improves the performance of a pretrained network on profile views without the need for ground-truth 3D shape information.

As illustrated in Fig. \ref{fig:method_overview}, the proposed technique starts from any existing method for single-image 3D face reconstruction and uses the 3D models produced by such method and their renderings at different viewpoints as data to fine-tune the original model.
The entire process requires neither additional 2D-3D ground-truth pairs, nor an additional deep-learning model for training.
Thus, the proposed bootstrap procedure is self-contained and works from any near-frontal face images, without annotations or 3D ground truth.

The main contribution of this work is the development of a self-supervised method that improves performance of a deep-learning model for single-image 3D face reconstruction on profile and near-profile views without the need to gather any additional 3D data, leading to better robustness to viewpoint variations on input images.
Furthermore, since only 2D images are required, the method can exploit images from large-scale, unconstrained, in-the-wild datasets \cite{KoestingerEtAl:AFLW:2011, Sagonas:2016:FIC:2949077.2949200, bulat2017far, LFWTech}, in contrast to the images present in 2D-3D datasets \cite{Bagdanov:2011:FHF:2072572.2072597, DBLP:journals/corr/ZhuLLSL15, BU4DFE, Cao:2014:FFE:2574216.2574355, Savran:2008:BDF:1505750.1505758, VWBS12}, which are typically of smaller scale and captured in more controlled environments.

\section{Related Work\label{sec:related_work}}
In this section we summarize state-of-art methods for single-image 3D face reconstruction.
We then highlight works related to self-supervised bootstrap training and self-training, and discuss their differences to our method.

\subsection{Single-Image 3D Face Reconstruction}
There is a large body of works on multi-view, image collection or video-based face reconstruction \cite{Beeler:2011:HPF:1964921.1964970, thies2016face, 7780824,Kemelmacher-Shlizerman:2011:FRW:2355573.2356525, kem:iccv13, DBLP:conf/eccv/SuwajanakornKS14}, but these methods are not the focus of the present work, which is concerned only with single-image face reconstruction.

A broad category of single-image face-reconstruction methods uses parametric models for representing the 3D shape of the faces.
The widely used 3D Morphable Model (3DMM) \cite{Blanz:1999:MMS:311535.311556, Romdhani2005Estimating3S} deploys an affine parametric model for face geometry, expression, and reflectance. 3DMMs represent the face geometry as a low-dimensional subspace obtained from the principal components of a set of high-resolution facial scans.
Original 3DMM methods worked by solving a non-linear optimization problem to fit a projection of a 3D model to 2D data  \cite{Blanz:1999:MMS:311535.311556, Romdhani2005Estimating3S, Huber2016AM3, DBLP:conf/iccv/RomdhaniV03, DBLP:journals/corr/JiangZDLL17}. 
However, recent advances in convolutional neural networks (CNNs) have made it possible to directly regress the 3DMM parameters from 2D images \cite{DBLP:journals/corr/TranHMM16,DBLP:journals/corr/ZhuLLSL15,DBLP:journals/corr/abs-1802-00542, DBLP:journals/corr/abs-1712-02859, Jourabloo2016LargePoseFA, DBLP:journals/corr/RichardsonSK16, DBLP:journals/corr/RichardsonSOK16, DBLP:journals/corr/KimZTTRT17}.
In Tran et al. \cite{DBLP:journals/corr/TranHMM16}, 3DMM parameters are directly regressed from pixel intensities, with training ground-truth generated from a robust multi-image face-reconstruction method \cite{Piotraschke2016Automated3F}.
In \cite{DBLP:journals/corr/ZhuLLSL15}, Zhu et al. proposed a method for 3DMM fitting under large pose variations, and created a large synthesized training set with profile views using a separate multi-feature 3DMM method \cite{Romdhani2005Estimating3S}.
However, the focus of their work is on 2D facial-landmark localization.
Furthermore, we do not require a separate method for generating varied-pose data, and our fine-tuning is done in a self-supervised bootstrapped manner, in which the network generates new training data to improve itself.
In \cite{DBLP:journals/corr/RichardsonSK16}, Richardson et al. used an iterative CNN for estimating the 3DMM parameters, and also use synthetically generated large-scale data for training.
However, they generate their synthetic data by sampling from a random normal distribution for the coefficients vector for 3DMM, whereas we target a specific region of the data domain for which it has been identified that the network has poor performance.
Recently, several methods \cite{DBLP:journals/corr/ChoyXGCS16, JacksonEtAl:FaceRecon:ICCV:2017} have attempted to predict 3D facial shapes using a volumetric rather than a parametric face model.
In \cite{JacksonEtAl:FaceRecon:ICCV:2017}, Jackson et al. used two stacked hourglass networks to directly predict the facial shape as an occupancy grid, and extracted a mesh using the marching cubes algorithm \cite{LorensenAndCline:MarchingCubes:1987}.
Another group of methods recently proposed \cite{DBLP:journals/corr/TewariZK0BPT17, DBLP:journals/corr/abs-1712-02859} used supervision purely on the 2D image domain, including landmarks and photometric loss, to achieve a self-supervised learning scheme for single-image face reconstruction.
In contrast to our approach, these methods must also rely on the existence of a parametric face model for training.

\subsection{Bootstrap Methods}
Self-training refers to the class of approaches that make predictions on unlabeled
data using a pretrained model, and use these predictions to further train the model itself.
Methods is this class has been previously applied to tasks such as image classification and object detection \cite{LiEtAl:Optimol:ICCV:2010, ChenEtAl:ExtractVisKnow:ICCV:2013, RosenbergEtAl:SemiSuperObjDet:WACV:2005, LaineEtAl:TempEnsembling:2016}.
Recently, Radosavovic et al. \cite{RadosavovicEtAl:DataDistilliation:ICCV:2017} adopted this approach and defined the notion of "data distillation," applying transformations to unlabelled images and grouping the predictions of a pretrained model on these transformed images as new labels.
These new labels along with the original manually labeled data serve as new data to fine-tune the pretrained model. This self-training scheme is related to the method herein proposed, with two crucial differences.
First, the transformations used in \cite{RadosavovicEtAl:DataDistilliation:ICCV:2017} are applied only to unlabelled images and are restricted to scaling and flipping, whereas in our method where we apply richer transformation to the predicted 3D shape.
Second, the transformation that we apply to the predicted 3D shape is guided towards identified regions of the data space where the pretrained model has poor performance.
Another work closely related to ours is \cite{DBLP:journals/corr/KimZTTRT17}, which adopts a parametric face model for simultaneously regressing all facial parameters from a single image.
That work also utilizes a bootstrap method via uniform re-sampling on the subspace of face-model parameters to generate new pairs of synthetic 2D images and parametric 3D shapes for iterative training.
Our method, on the other hand, is not bounded by a particular representation, i.e, we do not need to assume a parametric face model.
Moreover, we explicitly guide the network to improve its performance on regions of the 2D image domain where it does not perform well during the bootstrap process, instead of relying on random sampling as done in \cite{DBLP:journals/corr/KimZTTRT17}.

We need to select a baseline network to verify the effectiveness of our self-supervised bootstrap method in improving performance against large variations in pose.
Moreover, we want to demonstrate this improvement without the need to gather any additional 3D data or any modification to the original network architecture. 
To this end, we select the volumetric method in \cite{JacksonEtAl:FaceRecon:ICCV:2017} as the baseline network.
This approach produces high-quality results on input images that fall within the domain of the training data, i.e., near-frontal images with good illumination, but its performance drops as the characteristics of input images deviate from those of the original training domain, particularly for near-profile views, facial images with occlusions or atypical illumination.

\begin{figure}
    \centering
    \includegraphics[scale=0.5]{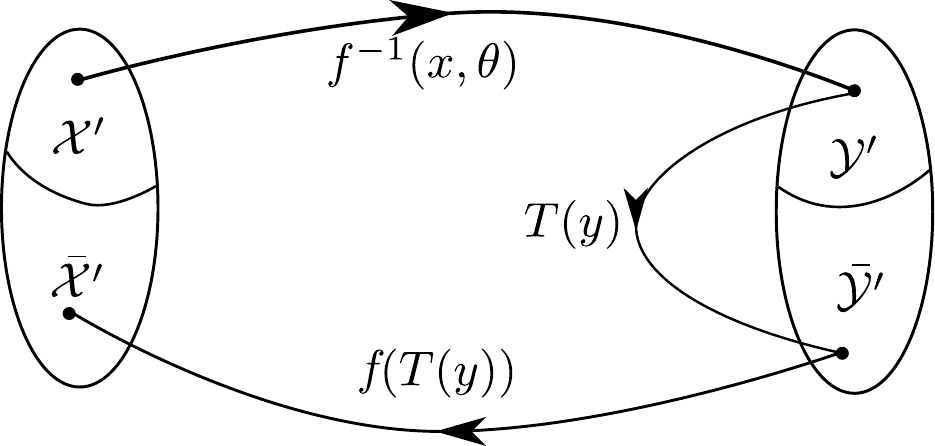}
    \caption{\label{fig:inverse_function}The diagram shows how, by starting with input data in a domain $\Domain'$ for which we have a good regression model $f^{-1}$, we synthesize good-quality data outside of $\Domain'$.}
\end{figure}

\section{Description of Method\label{sec:method}}

Consider an inverse problem requiring the estimation of parameters $\param\in\Param$ of a function
$f^{-1} : \Domain\times\Param\mapsto \Image$.
Assume that the direct problem of estimating $x\in\Domain$ given $y\in\Image$ is ``simple'' in some reasonable sense, and can be solved through a known function $f:\Image\mapsto\Domain$.
The core idea of the method proposed in this paper is to apply the inverse function $f^{-1}$ to a subset of $\Domain'\subset\Domain$ such that $\hat{y} = f^{-1}(x,\param)$ is a ``good'' estimator of a true value $y$ for $x\in\Domain'$, apply to $\hat{y}$ a transformation $T:\Image\mapsto\Image$, and utilize the new pair $(f\circ T(\hat{y}), T(\hat{y}))$ as input data for the refinement of the current estimate of \param.
The function $T$ should be selected as to move $\hat{x} = f\circ T(\hat{y})\in\Domain$ towards regions of the domain \Domain for which the original estimate \param of the parameters of $f^{-1}$ provides a poor approximation for the actual map between \Domain and \Image.
This procedure is illustrated in Fig.~\ref{fig:inverse_function}

When applying this framework to a deep model, the function $f^{-1}$ corresponds to a particular model, and \param corresponds to an initial setting of its weights.
Input data $x'\in\Domain'$ is provided to the network, producing outputs $y'\in\Image'\subset\Image$.
Through appropriate transformations $f$ and $T$, new pairs $(f\circ T(y'), T(y'))$ are produced, and set aside as new training data.
Note that these training pairs have been produced without knowledge of the ground-truth value corresponding to the inputs in $\Domain'$.
The problem of 3D estimation from a single image is particularly amenable to this approach, since it is a difficult inverse problem with a corresponding direct problem defined by a relatively simple function $f$.

In the context of this paper, we identify $\mathcal{X}$ and $\mathcal{Y}$ with 2D and 3D domains.
The function $f^{-1}$ corresponds to a deep-learning model that performs 3D face reconstruction from a single image, and the parameter $\param$ corresponds to the weights of that model.
The function $f$ is the corresponding direct problem, i.e., projection and rendering, and the function $T$ corresponds to a rigid-body transformation.

\subsection{Selection of Data Domain $\Domain'$}

The application of the proposed bootstrap method requires the identification of subsets of the input domain \Domain for which the network has ``good'' and ``bad'' performances.
One possible way to make this identification is to define a partition of $\cup_{i=1}^{N}\Domain_{i}$ of \Domain, and evaluate the performance of the network on each subset $\Domain_{i}$.
This approach suffers from two difficulties: first, it requires the definition of criteria for the partition of \Domain; second, it relies on the availability of ground truth over the full domain \Domain.
Instead, the approach proposed here is to assume prior knowledge of a subset of $\Domain$ for which the network is expected to have ``good'' performance, and use the bootstrap method itself to identify subsets of $\Domain$ with sub-par performance.

This procedure is problem- and data-dependent; for the specific problem of single-image face reconstruction, it has been well-documented, as discussed in Section \ref{sec:related_work}, that most methods underperform for profile or near-profile views.
The converse of this observation leads to the assumption that a reasonable choice for $\Domain'$ is the subset of \Domain consisting of fronto-parallel views.
We can then apply to the output of the network under inputs in $\Domain'$ a transformation $T$ that rotates the reconstructed 3D models away from the viewing direction of the camera.
These transformed models are then rendered, and fed back to the network for reconstruction.
Regions of the input domain are therefore indexed by the parameters of the transformation $T$, and those regions for which the model performs poorly can be identified by comparison between two 3D models: a higher-quality model obtained by applying $T$ to the 3D model obtained from an input in $\Domain'$, and the 3D model directly obtained from the reconstruction of the rendering of the transformed models.

\begin{figure}[tbh]
\centering
\includegraphics[width=\textwidth]{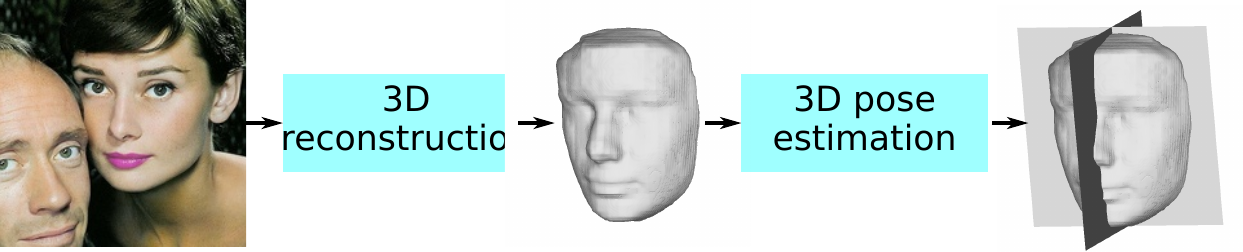}
	\caption{\label{fig:reconstruction_bilateral_symmetry} Typical output of VRN-Unguided network \cite{JacksonEtAl:FaceRecon:ICCV:2017}, converted to mesh from 3D volume; Extraction of bilateral symmetry and ``gaze'' planes of the face reconstructed.}
\end{figure}

\subsection{Procedure for Self-Supervised Bootstrap\label{sec:procedure_bootstrap}}

The procedure for the proposed self supervised bootstrap method has four steps: \begin{enumerate*}[label=(\roman*)] \item 3D face reconstruction of an existing model, \item application of rigid-body transformations, \item rendering, and \item bootstrap fine-tuning \end{enumerate*}.
\begin{figure*}
  \tabcolsep=0pt
	\begin{tabu} to \textwidth {X[0.5,c,m]*7{X[1,c,m]}}
	 & original & $+20^\circ$ & $+30^\circ$ & $+40^\circ$ & $+50^\circ$ & $+60^\circ$ & $+70^\circ$\\ 
	$-20^\circ$ &
	\includegraphics[width=0.125\textwidth]{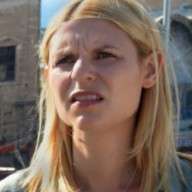}&
	\includegraphics[width=0.125\textwidth]{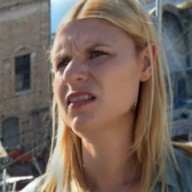}&
	\includegraphics[width=0.125\textwidth]{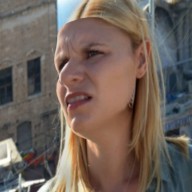}&
	\includegraphics[width=0.125\textwidth]{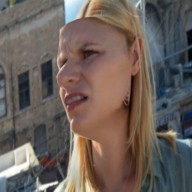}&
	\includegraphics[width=0.125\textwidth]{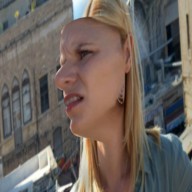}&
	\includegraphics[width=0.125\textwidth]{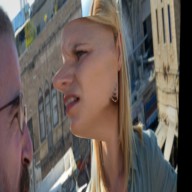}&
	\includegraphics[width=0.125\textwidth]{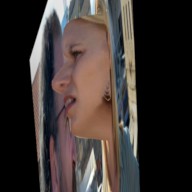}\\
	\rotatebox[origin=c]{90}{original} &
	\includegraphics[width=0.125\textwidth]{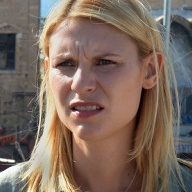}&
	\includegraphics[width=0.125\textwidth]{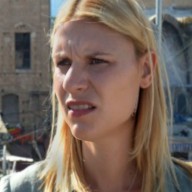}&
	\includegraphics[width=0.125\textwidth]{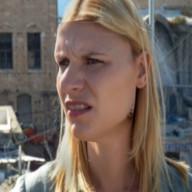}&
	\includegraphics[width=0.125\textwidth]{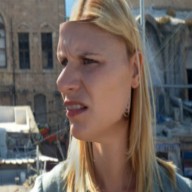}&
	\includegraphics[width=0.125\textwidth]{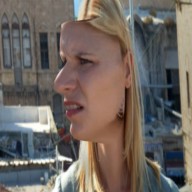}&
	\includegraphics[width=0.125\textwidth]{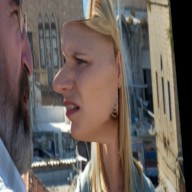}&
	\includegraphics[width=0.125\textwidth]{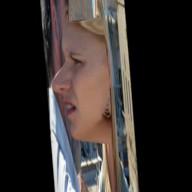}\\
	$+20^\circ$ &
	\includegraphics[width=0.125\textwidth]{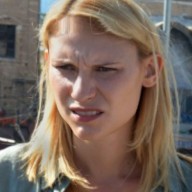}&
	\includegraphics[width=0.125\textwidth]{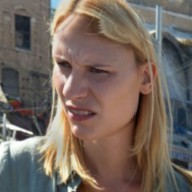}&
	\includegraphics[width=0.125\textwidth]{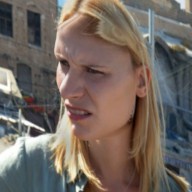}&
	\includegraphics[width=0.125\textwidth]{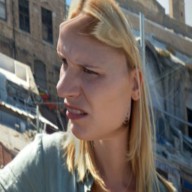}&
	\includegraphics[width=0.125\textwidth]{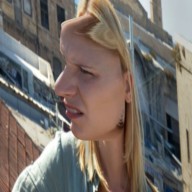}&
	\includegraphics[width=0.125\textwidth]{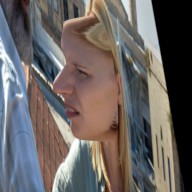}&
	\includegraphics[width=0.125\textwidth]{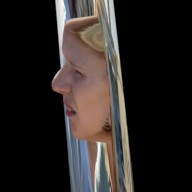}
  	\end{tabu}
    \caption{\label{fig:rendering} Renderings of the input image from different viewing angles.}
\end{figure*}

\subsubsection{3D face reconstruction}
We apply our self-supervised bootstrap technique to the \emph{volumetric regression network} (VRN) method \cite{JacksonEtAl:FaceRecon:ICCV:2017}. However, we highlight that the proposed technique can be applied to any other data-driven single-image 3D-reconstruction method. VRN directly regresses a 3D occupancy volume from a single 2D input image, which is then converted into a mesh by the marching-cubes algorithm \cite{LorensenAndCline:MarchingCubes:1987}. A typical output of VRN-Unguided is shown in \figref{fig:reconstruction_bilateral_symmetry}.

\subsubsection{Application of rigid-body transformations\label{sec:rigid_body_trans}}
Starting with high-quality reconstructed 3D models from fronto-parallel views, we transform them according to a rigid-body motion, steering the volumes towards profile or near-profile views, where a given network does not perform as well.
In order to specify the pose of the new views, it is first necessary to know what are the poses of the input faces with respect to the camera.
We represent the current pose of a face by describing its plane of bilateral symmetry and its ``backplane,'' which is a plane orthogonal to both the face's plane of symmetry and its gaze direction. The gaze direction is defined as the direction pointed at by the subject's nose.
Estimation of the bilateral symmetry plane is achieved by observing that the plane normal is an eigenvector of the sample covariance matrix $\hat{\makebold{\Sigma}}$ of the vertices in a mesh representation of the face.
Moreover, due to the nature of the VRN reconstruction, which produces ``shallow'' faces akin to a face mask rather than a full skull, the other two eigenvectors of $\hat{\makebold{\Sigma}}$ correspond to the gaze direction and the ``vertical'' direction of the face, pointing towards the top of the subject's head.
The bilateral symmetry and backplane plane of the 3D face are shown in \figref{fig:reconstruction_bilateral_symmetry}. Detailed derivations of the bilateral symmetry plane and the backplane are provided in the supplementary material.
In order to produce more realistic renderings, it is important to add a background to the images.
We use the original image as the background, textured mapped onto the backplane.

\subsubsection{Rendering of new viewpoints\label{sec:rendering}}
To produce realistic renderings we use an emissive illumination model, where the material of each vertex has no reflectance component and behaves instead as a light source.
We generate novel viewpoints of the original model by rotating it around axes $y$ (yaw) and $x$ (pitch), in increments of $10^{\circ}$.
The model is rotated away from the bilateral symmetry plane, up to the maximal angle such that the gaze direction does not exceed $90^{\circ}$ with respect to the camera viewing direction.
Finally, we constrain the rotation angle around $x$ to the interval $[-20^{\circ},20^{\circ}]$, to avoid extreme top or bottom views.
An example of this procedure is show in \figref{fig:rendering}.

\subsubsection{Bootstrap fine-tuning\label{sec:bootstrap_finetune}}
The final step of the proposed self-supervised bootstrap approach is to use the rendered 2D images from step (iii) and their 3D counterparts, which are generated with the pre-trained network in step (i) and modified through step (ii), as additional data to fine-tune the original network for face reconstruction. In this paper, we focus on applying the fine-tuning on the network architecture of VRN-Unguided. Furthermore, it is essential to note that there are no additional changes to the original network architecture or loss function which allows the flexibility of applying this same bootstrap approach to any other deep-learning architecture for single-image face reconstruction. In addition, as there is no requirement to gather 3D ground-truth for the self-supervised bootstrap procedure, we can use any in-the-wild 2D face images to improve the given pretrained network. This capability of the bootstrap method allows for infinite amount of fine-tuning data and promising results are shown in the experimental section \ref{sec:experiments} with bootstrapping using the well-known 2D face image dataset LS3D-W \cite{bulat2017far}.

\section{Experimental Results\label{sec:experiments}}
We perform three major experiments.
First, we analyze the performance of a state-of-the-art model for 3D face reconstruction against variations in the pose of the input images.
The second experiment demonstrates the effectiveness of the proposed method in increasing the performance of a state-of-art pretrained network.
Reconstruction quality of the model before and after applying the bootstrap method is evaluated on a well-known benchmark dataset for single image face reconstruction.
The third experiment further verifies the effectiveness of the proposed self-supervised bootstrap method by evaluating the model performance with and without bootstrapping on a benchmark dataset with a more uniform pose distribution.

\subsection{Datasets and Evaluation Protocol}
Since the proposed self-supervised bootstrap method does not require additional 3D ground-truth shape data, we can utilize any dataset of face images in the bootstrap procedure. For all three experiments conducted in this work, \textbf{LS3D-W} \cite{bulat2017far}, a large facial 2D image dataset containing \texttildelow 200K unconstrained face images in the wild was used for bootstrapping. For the first experiment of analysis on reconstruction robustness against pose variation, we used the \textbf{300W-Test-3D} \cite{bulat2017far} dataset containing 600 unconstrained 2D face images. For quantitative evaluation of reconstruction accuracy, we used the dataset \textbf{AFLW2000-3D} \cite{DBLP:journals/corr/ZhuLLSL15}, which contains 2000 pairs of facial 2D images to 3D ground truth shapes and the \textbf{MICC} \cite{Bagdanov:2011:FHF:2072572.2072597} dataset which has large pose variation for reconstruction quality evaluation. It is noted that during the bootstrapping procedure using the 2D image dataset LS3D-W, we exclude its subset of 300W-Testset-3D and AFLW2000-Reannotated which are partly used as testing images in the first two experiments described above even though they do not have associated 3D shape counterpart within the LS3D-W dataset.

\subsubsection{Error metric}
To compare a ground-truth 3D face shape against a predicted shape we use the normalized mean error (NME) as defined in \cite{JacksonEtAl:FaceRecon:ICCV:2017}, i.e., the average $L_{2}$ distance between closest points on the meshes normalized by the 3D outer inter-ocular distance $\text{d}$
\begin{align}
NME = &\frac{1}{N} \sum_{i = 1 }^{N} \frac{||\boldsymbol{\hat{x_{i}}} - \boldsymbol{x_{i}}||_{2}} {\text{d}}\label{eq:nme},
\end{align} where $N$ is the number of vertices used for estimating the $L_{2}$ distance on each mesh reconstructed, $ \boldsymbol{\hat{x_{i}}}$ and $  \boldsymbol{x_{i}} $ are the estimated and ground-truth vertex locations.
This measure can be interpreted as the percentage of error distance over the 3D outer inter-ocular distance. 

\subsubsection{Implementation details}
To demonstrate the effectiveness of the proposed method, we take the architecture of VRN-Unguided without modification and perform the bootstrap fine-tuning as described in section \ref{sec:procedure_bootstrap}. In all three experiments, 
the bootstrap fine-tuning is performed on the 2D face-image dataset LS3D-W \cite{bulat2017far}. Rigid-body transformations applied in the bootstrap process are yaw rotations around the y-axis of the camera coordinate system of the input face images. For network fine-tuning, we adopt an initial learning rate of $10^{-6}$, batch size of 64 and use Adam optimizer \cite{DBLP:journals/corr/KingmaB14}. We decrease the learning rate with a factor of 0.5
every 5 epochs. For hyper-parameter tuning, we split the new pairs of 2D image to 3D shape bootstrapped from LS3D-W to training and validation set following a 90\% and 10\% split. The final model picked for experiment results shown in this section is at epoch 10.

\subsubsection{Overview of results}
\begin{figure}[htb]
    \centering
    \includegraphics[width=\linewidth]{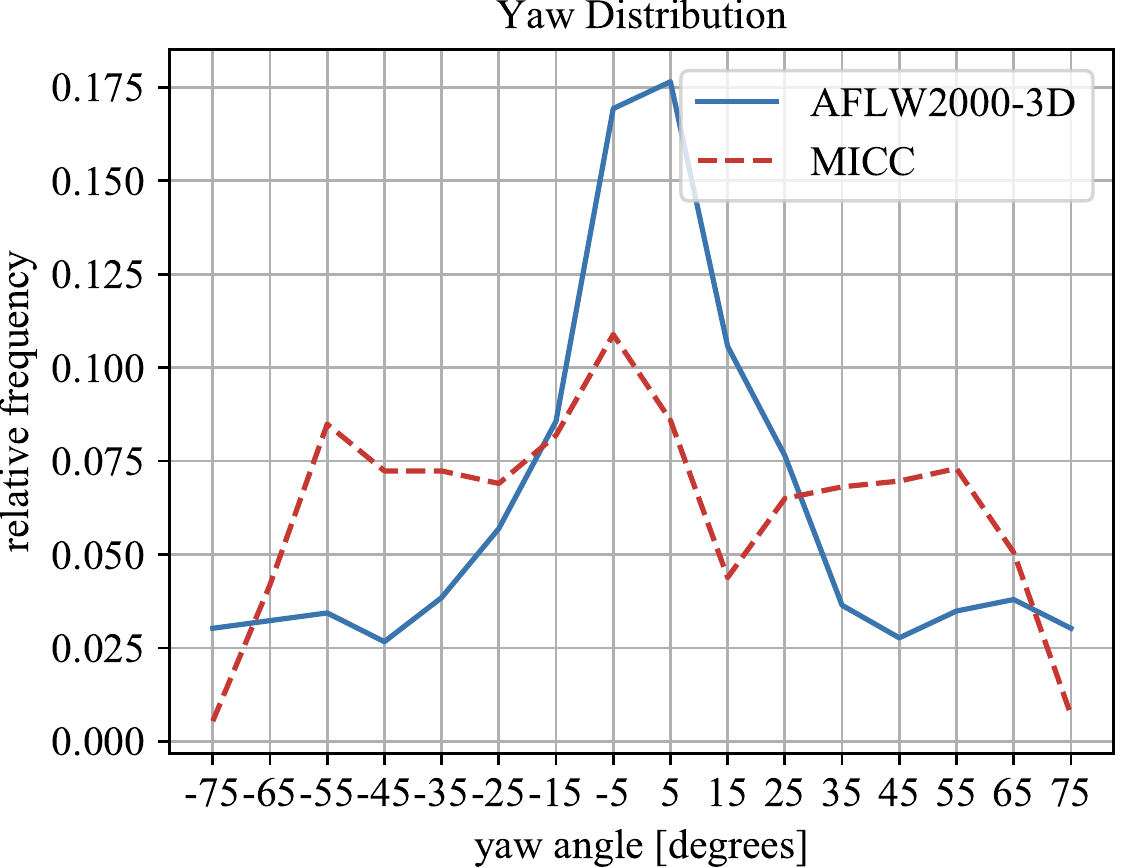}
    \caption{\label{fig:yaw_dist} Distribution of yaw angles in two evaluation datasets. The distribution is more uniform for the MICC dataset than for the AFLW2000 dataset.}
\end{figure}

Table \ref{tbl:quantitative_performance} summarizes the quantitative comparison of methods with and without the self-supervised bootstrap approach on the aforementioned datasets for reconstruction quality evaluation. We compare the performance of models bootstrapped with different amount of rigid body transformations of, $\pm20^{\circ}\text{\!\!,}$ $\pm40^{\circ}\text{\!\!,}$ $\pm60^{\circ}$ in yaw, $\pm20^{\circ}\text{\!\!,}$ $\pm40^{\circ}$ in yaw and $\pm20^{\circ}$ only, respectively.
It can be seen that bootstrapping with yaw rotations of $\pm20^{\circ}\text{\!\!,}$ $\pm40^{\circ}\text{\!\!,}$ $\pm60^{\circ}$ results in the lowest reconstruction error.
The overall relative reconstruction quality improves by 4.0\% for VRN-Unguided on AFLW2000-3D dataset and 7.7\% on the MICC dataset which has larger pose variation and a much more uniform distribution over frontal to profile views, as shown in Fig. \ref{fig:yaw_dist}.
We summarize results of the three experiments as follows:
\begin{itemize}
    \item The benchmarked state-of-art volumetric regression model for single image 3D face reconstruction has a performance bias towards fronto-parallel viewpoints and its performance deteriorates as the camera angle moves away from fronto-parallel views.
    \item The model with the self-supervised bootstrap procedure outperforms the original pretrained model in reconstruction quality for all datasets evaluated, especially for faces with large rotation angle in yaw.
    \item The self-supervised bootstrap procedure increases the network's robustness against pose variations, resulting in a more uniform performance across viewing angles.
    Furthermore, the greater the rigid-body rotations applied during bootstrapping, the more robust the fine-tuned model becomes against large pose variation.
\end{itemize}

\begin{table}[bth]
\setlength{\tabcolsep}{2pt}
\centering
    \begin{tabular}{>{\centering\arraybackslash}p{4.0cm}>{\centering\arraybackslash}p{2.3cm}>{\centering\arraybackslash}p{1.1cm}}
    	VRN variant & AFLW2000-3D & MICC\\ [0.5ex] 
    	\hline\hline
    	original & 1.98\% & 2.72\% \\ 
    	\hline
    	btstrppd ($\pm20^{\circ}$ yaw) & 2.02\% & 2.67\%\\
    	\hline
    	btstrppd ($\pm20^{\circ}\text{\!\!,}$ $\pm40^{\circ}$) & 1.93\% & 2.55\% \\
    	\hline
    	btstrppd ($\pm20^{\circ}\text{\!\!,}$ $\pm40^{\circ}\text{\!\!,}$ $\pm60^{\circ}$) & \textbf{1.90\%} & \textbf{2.51}\% \\
    	\hline
    \end{tabular}
    \caption{\label{tbl:quantitative_performance} NME error on AFLW2000-3D and MICC datasets.\vspace*{-3ex}}
\end{table}

\begin{figure}[tb]
        \includegraphics[width=1.0\linewidth]{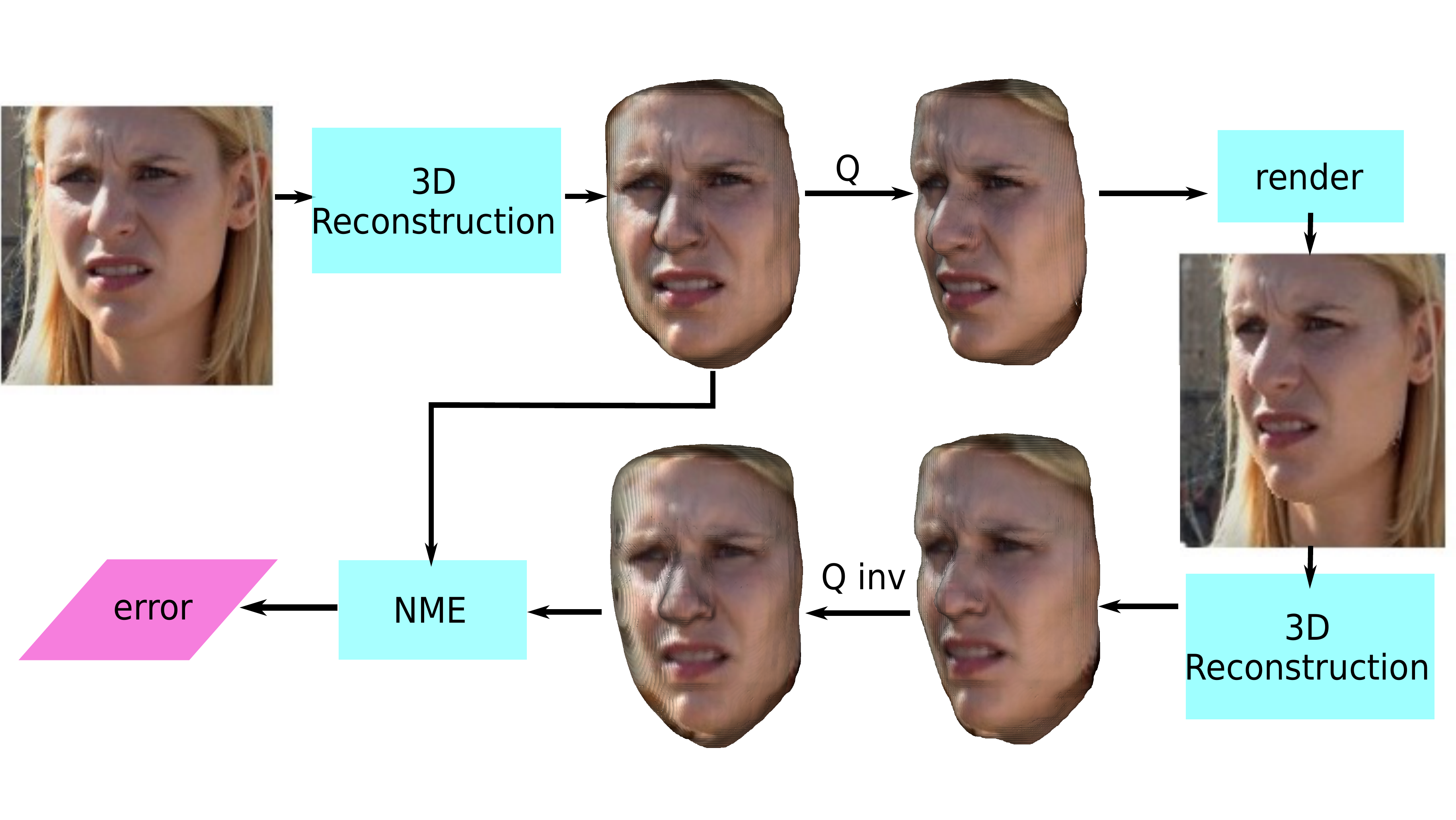} 
  		\caption{ \label{fig:self_recon_experiment} Experimental design for analysis of robustness of VRN against pose variation. The two 3D-reconstruction stages use exactly the same deep-learning model.}
\end{figure}

\begin{figure}[htb]
    \centering
    \includegraphics[width = 1.0\linewidth]{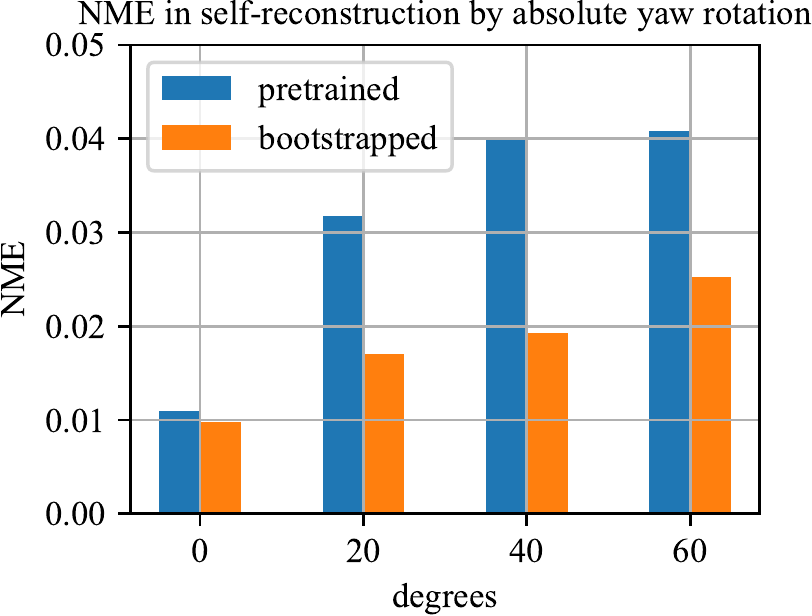}
    \caption{ \label{fig:self_recon_error}
		Error in reconstruction from rendered images at a given pose compared to the original 3D shape from near-frontal views.
			Blue bars show results before applying the proposed method; orange bars show results after applying bootstrap with $\pm20^{\circ}\text{\!\!,}$ $\pm40^{\circ}\text{\!\!,}$ $\pm60^{\circ}$ rigid body rotation in yaw.}
\end{figure}

\begin{figure}[tbh]
        \includegraphics[width=\linewidth]{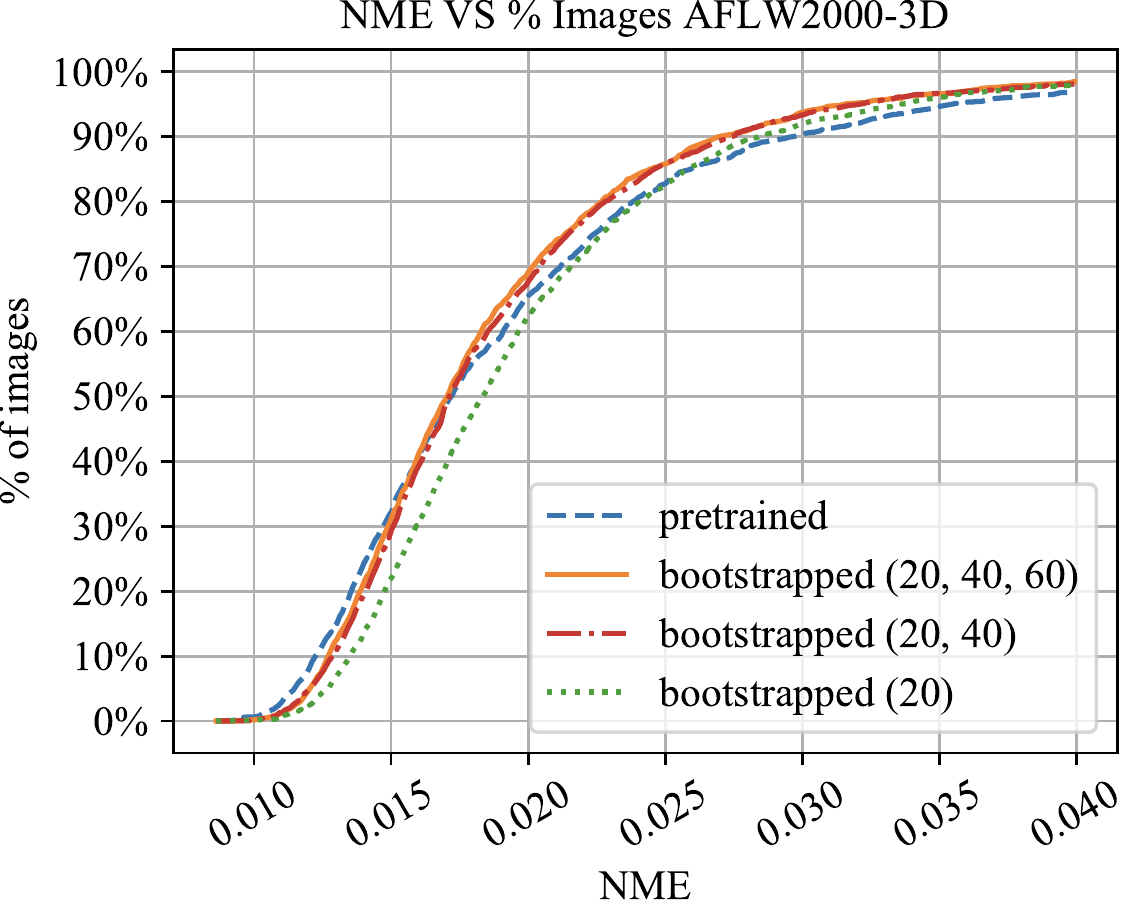} 
  		\caption{ \label{fig:nme_percent_AFLW2000} NME vs \% Images on AFLW2000-3D dataset.}
\end{figure}
  
\subsection{Reconstruction Robustness vs.\ Pose Variation}

As shown in Fig. \ref{fig:self_recon_experiment}, we first explore the robustness of 3D reconstructions of the state-of-art VRN method against pose variations.
We conduct the experiment on VRN-Unguided architecture as this is the only pretrained model provided in \cite{JacksonEtAl:FaceRecon:ICCV:2017}.
To analyze the quality and robustness of the 3D reconstruction of the VRN model for inputs with different viewing angles, we first reconstruct the face geometry of near fronto-parallel inputs.
We then apply a known rigid-body transformation to the 3D models thus produced, and generate a realistic rendering of the transformed 3D face.
Finally, another 3D face model is reconstructed using the rendered image, and we apply to that model the inverse of previously defined rigid-body transformation.
We then compare the result of this transformation with the original reconstruction result using the NME metric.
If the reconstruction method is robust against pose variations, a flat distribution of small errors ought to be observed, due solely to rendering artifacts introduced by the pipeline.
In this experiment, we use the 600 test face images from 300W-Testset-3D set, which is a subset of the larger LS3D-W dataset.
Results of this experiment are shown in Fig. \ref{fig:self_recon_error}.
It can be observed that after applying the self supervised bootstrap method, the model exhibits a more uniform error distribution, which indicates an increased capacity of the network in dealing with non-frontal views.

\begin{figure}[tbh]
        \includegraphics[width=\linewidth]{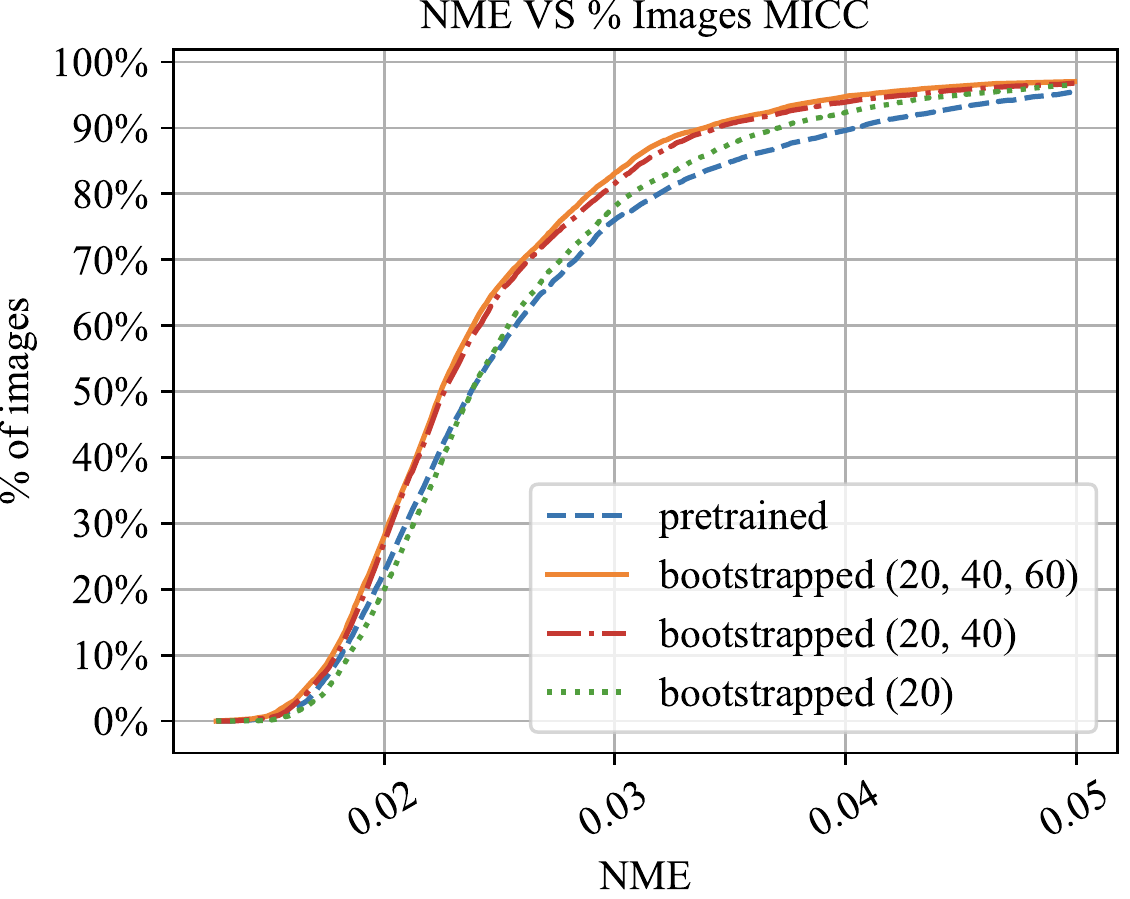} 
  		\caption{ \label{fig:nme_percent_MICC} NME vs \% Images on MICC dataset}
\end{figure}

\subsection{Evaluation on AFLW2000-3D Dataset}
\begin{figure*}[bh]
  \tabcolsep=0pt
	\begin{tabu} to \textwidth {X[0.65,p,m]*6{X[1,c,m]}}
    \centering 2D image&
	\includegraphics[width=0.1538\textwidth]{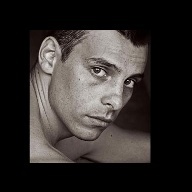}&
	\includegraphics[width=0.1538\textwidth]{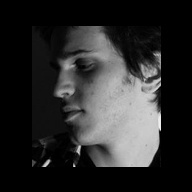}&
	\includegraphics[width=0.1538\textwidth]{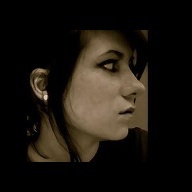}&
	\includegraphics[width=0.1538\textwidth]{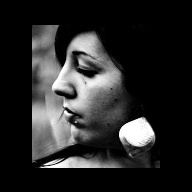}&
	\includegraphics[width=0.1538\textwidth]{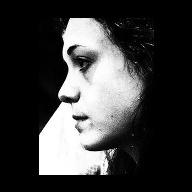}&
	\includegraphics[width=0.1538\textwidth]{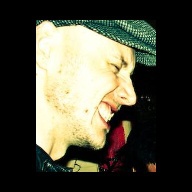}\\
    \centering without bootstrap &
	\includegraphics[width=0.1538\textwidth]{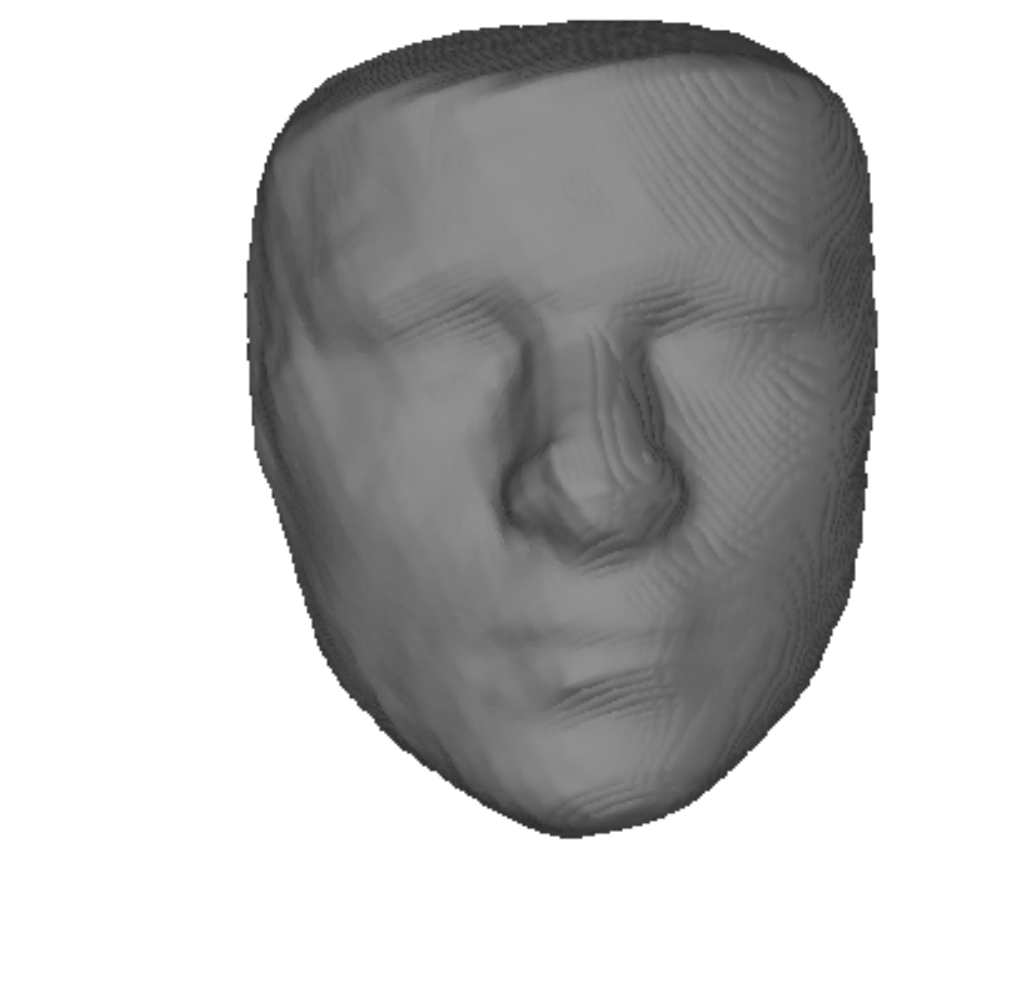}&
	\includegraphics[width=0.1538\textwidth]{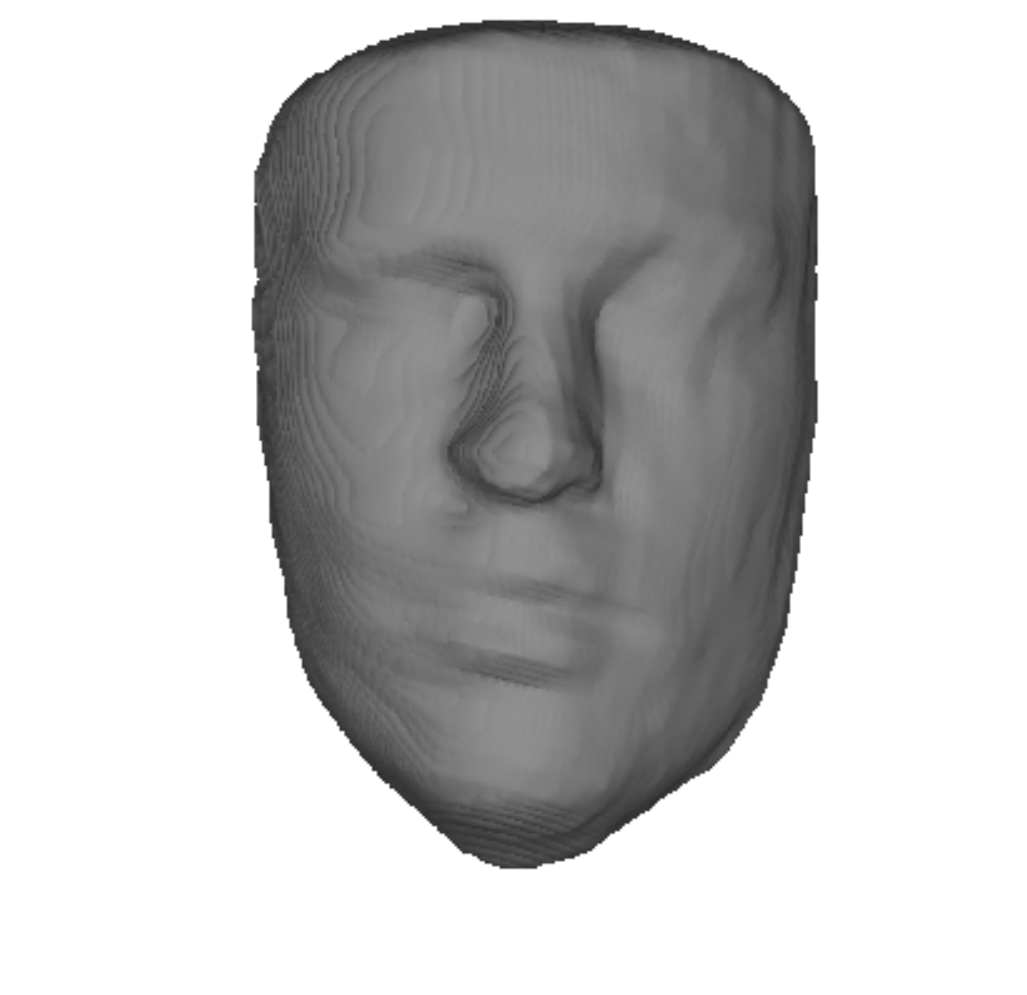}&
	\includegraphics[width=0.1538\textwidth]{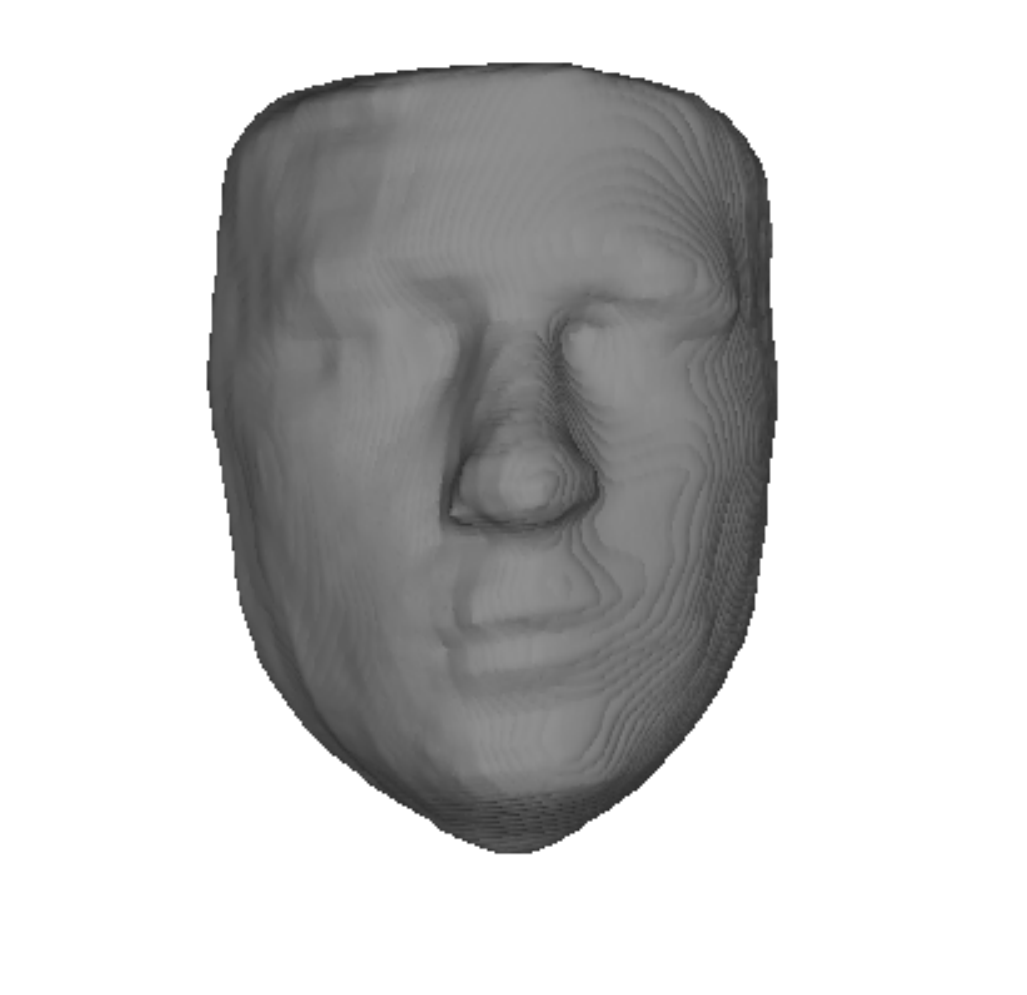}&
	\includegraphics[width=0.1538\textwidth]{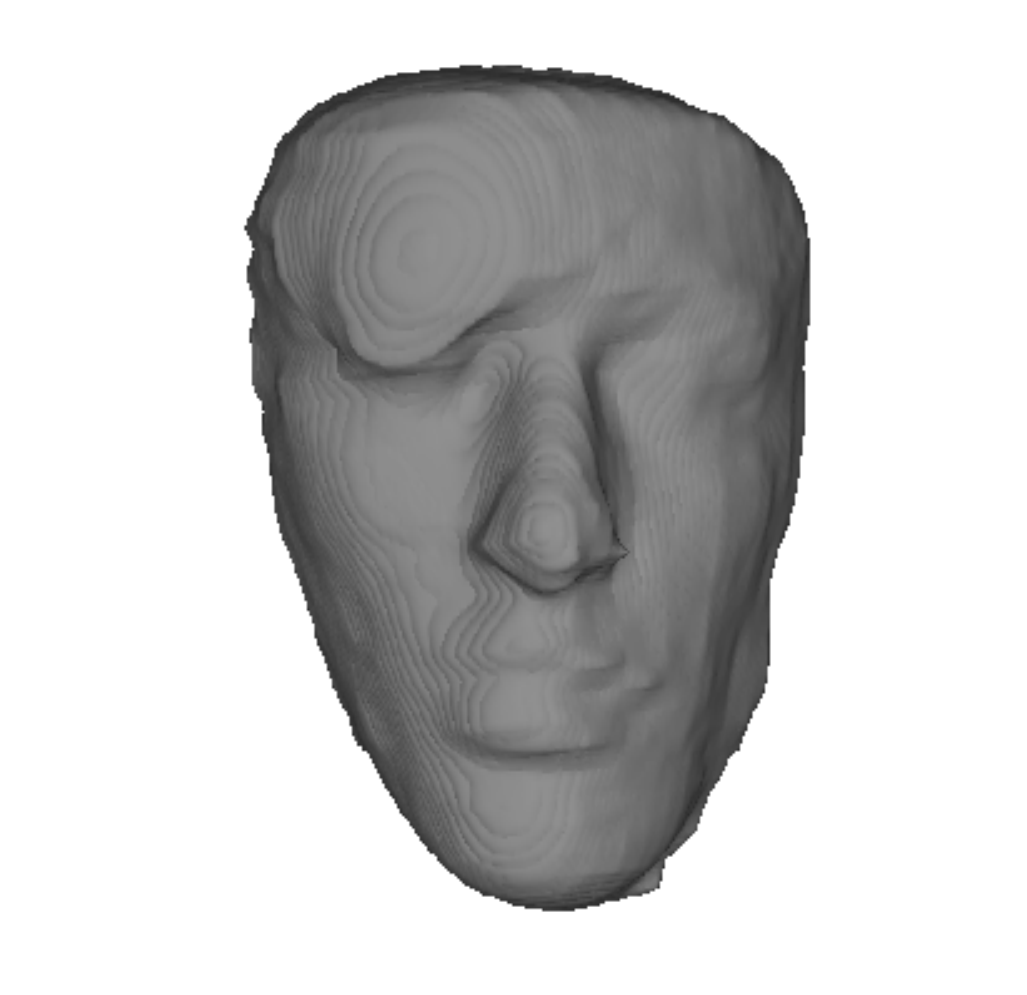}&
	\includegraphics[width=0.1538\textwidth]{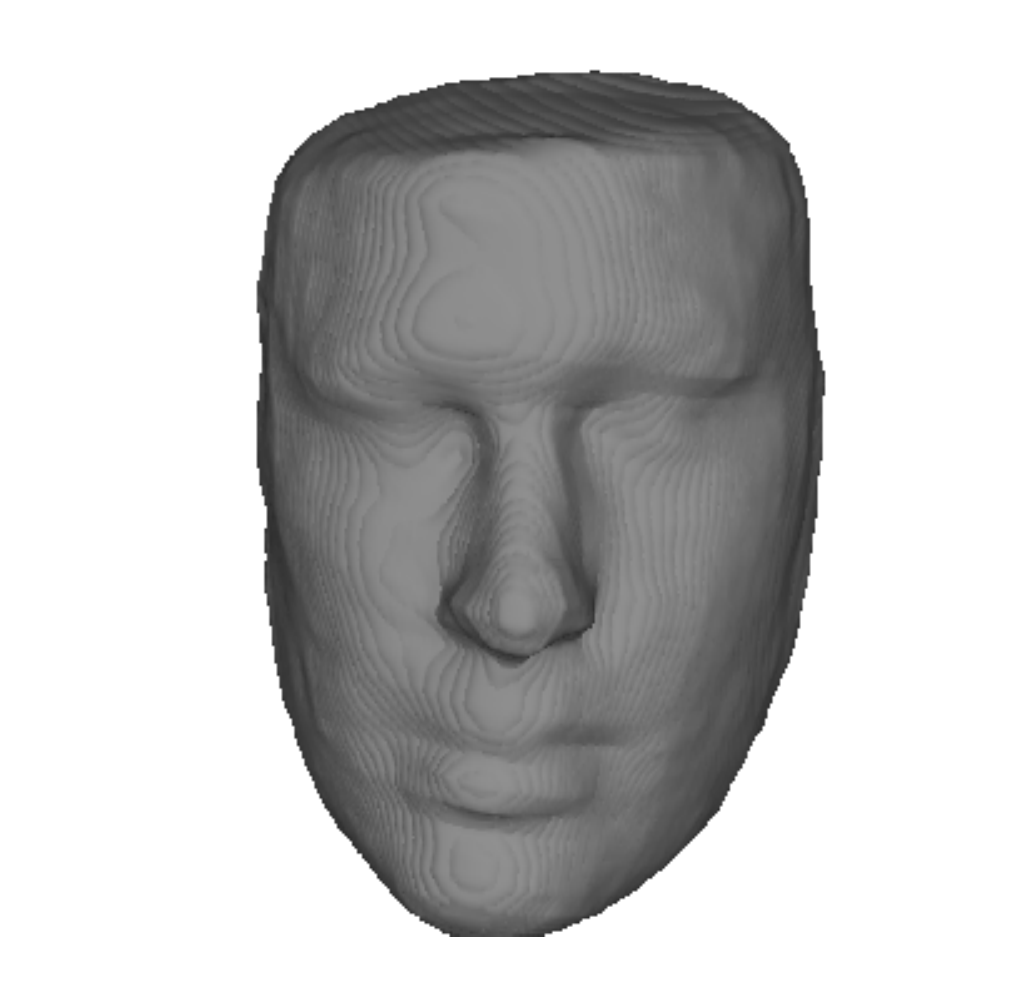}&
	\includegraphics[width=0.1538\textwidth]{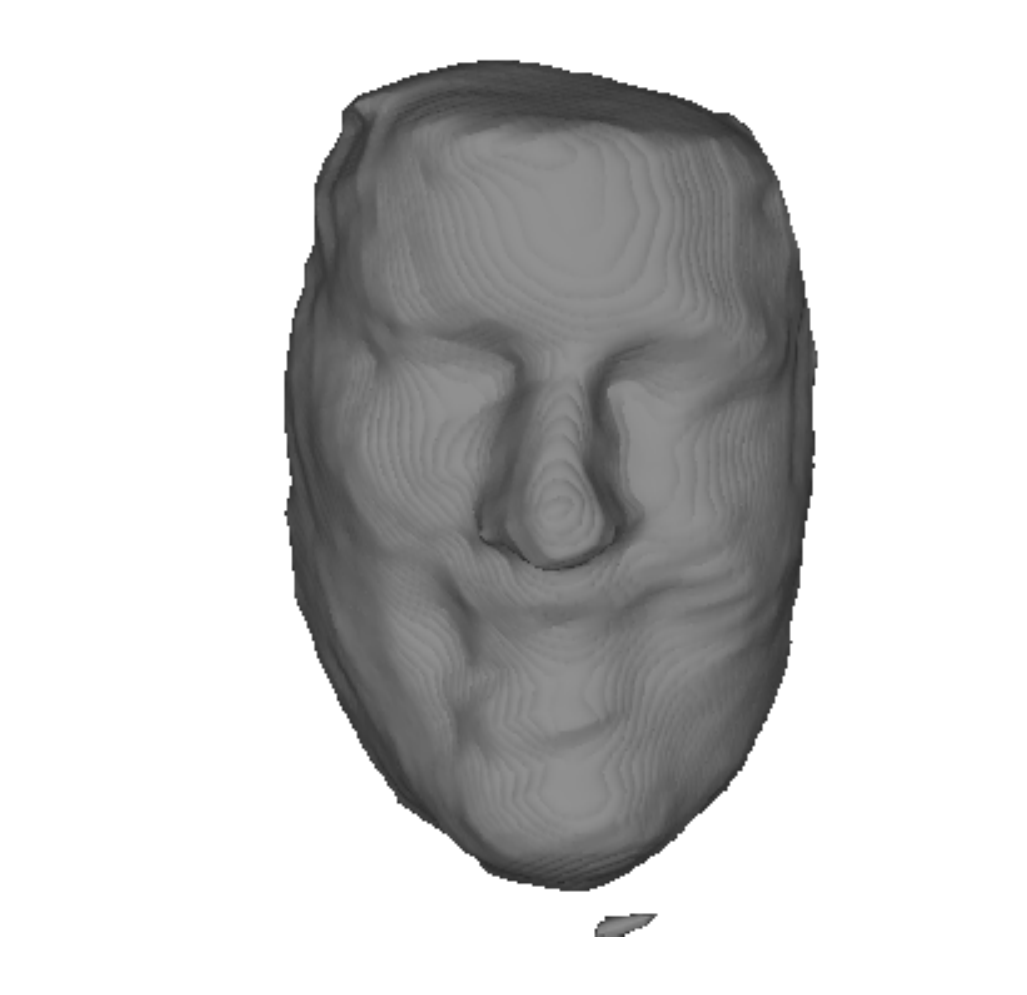}
	\\
    \centering with bootstrap &
	\includegraphics[width=0.1538\textwidth]{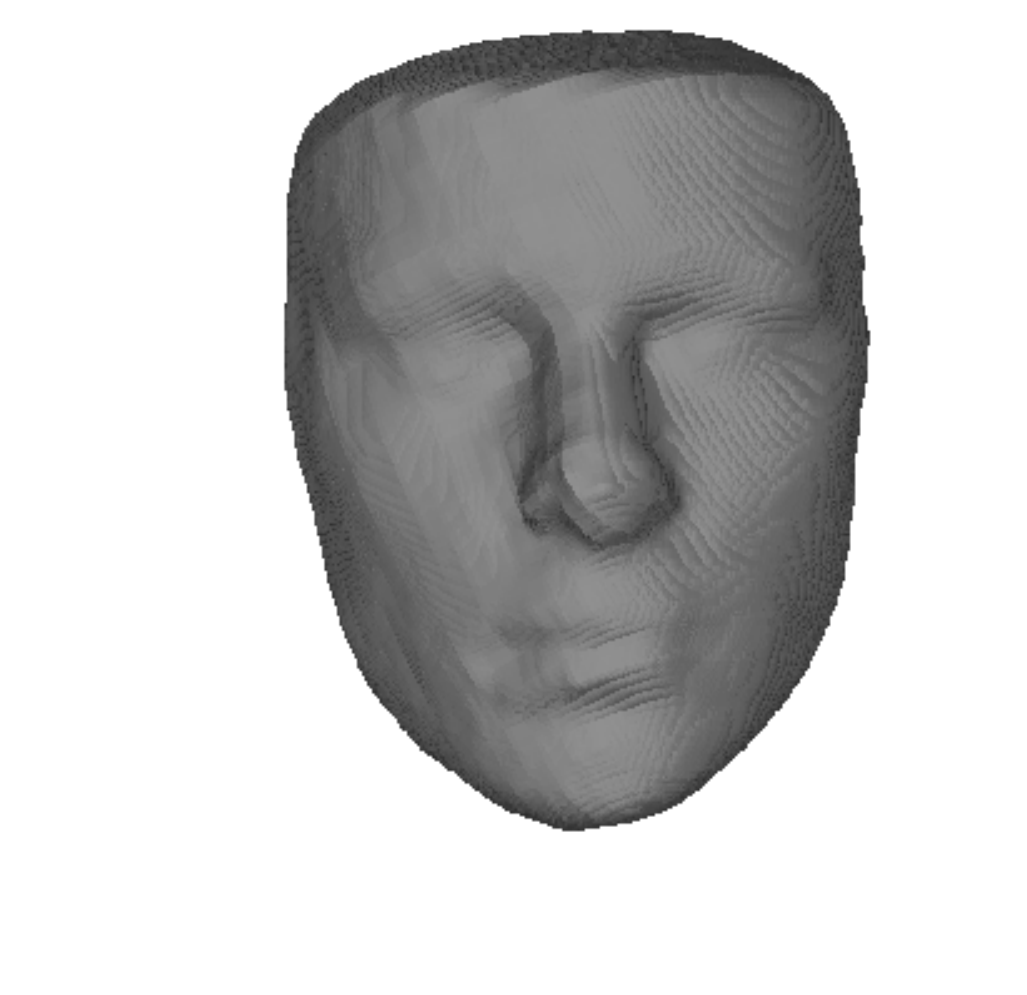}&
	\includegraphics[width=0.1538\textwidth]{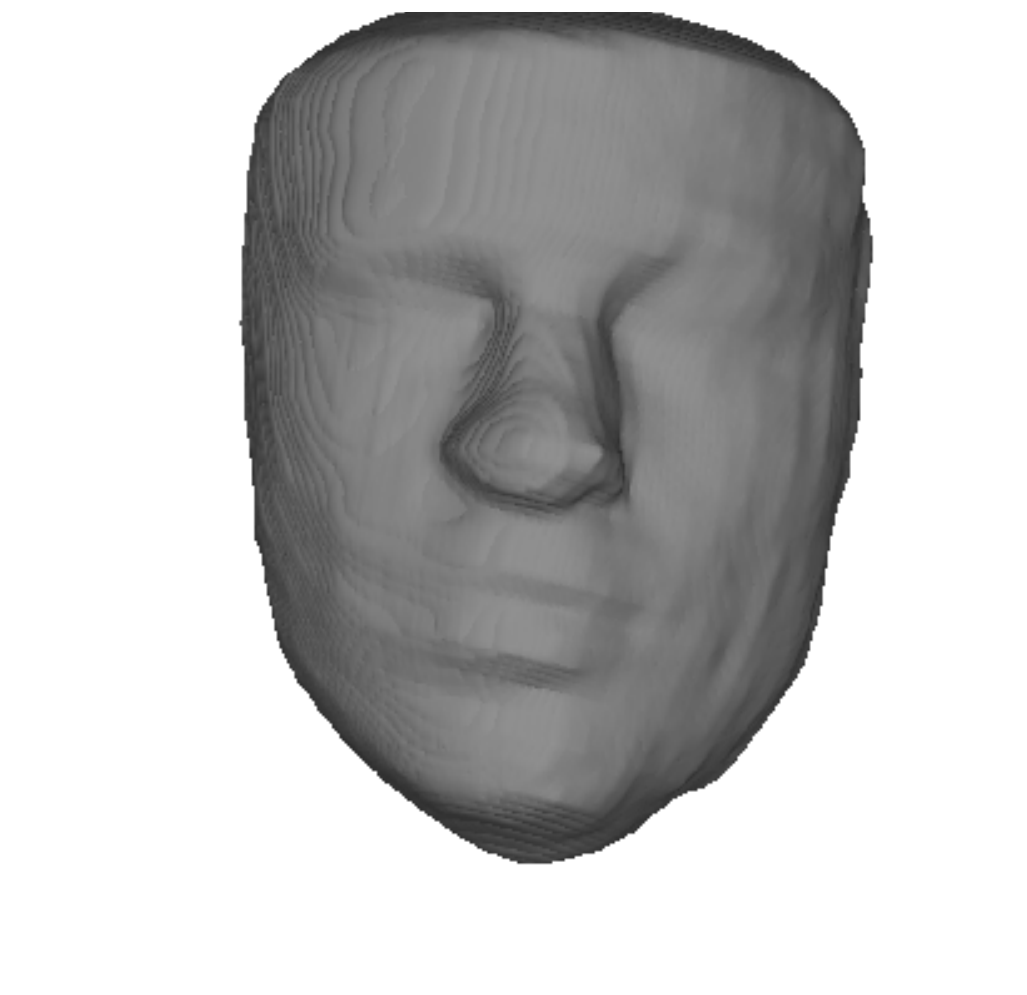}&
	\includegraphics[width=0.1538\textwidth]{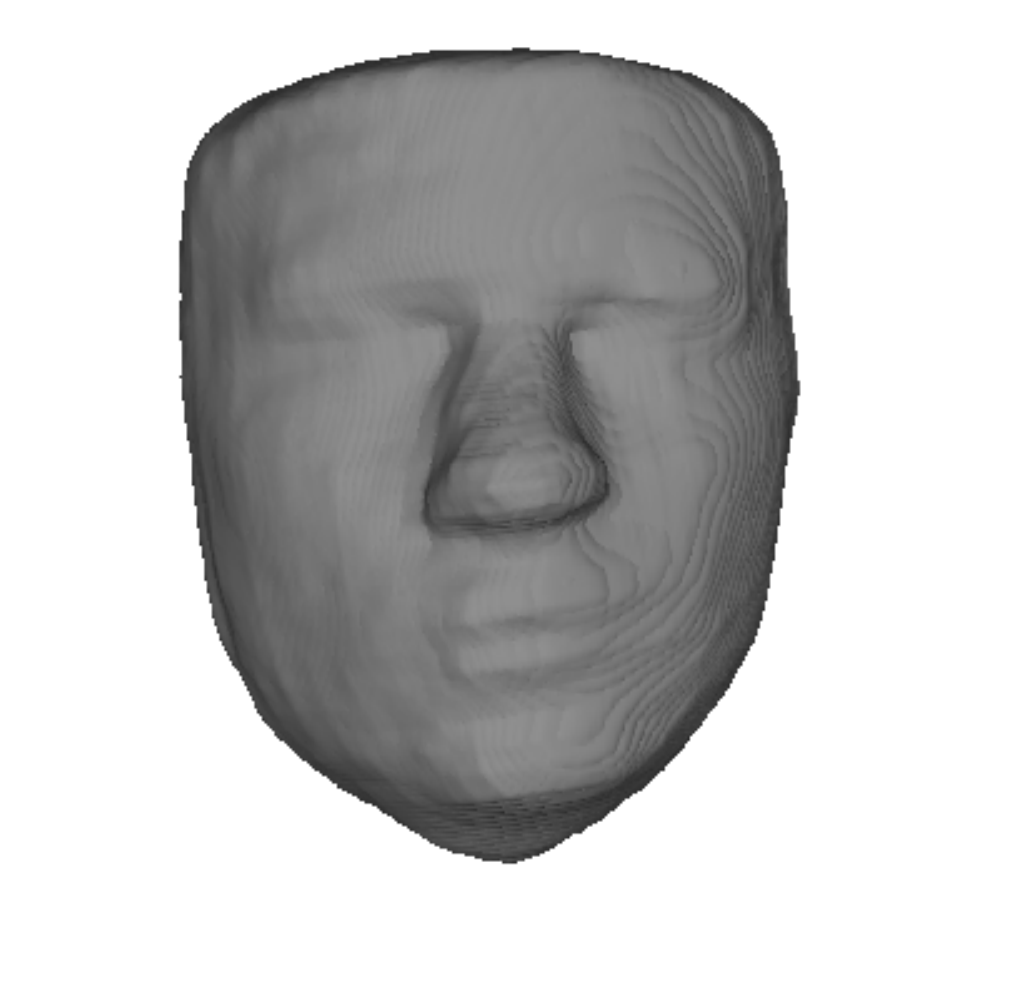}&
	\includegraphics[width=0.1538\textwidth]{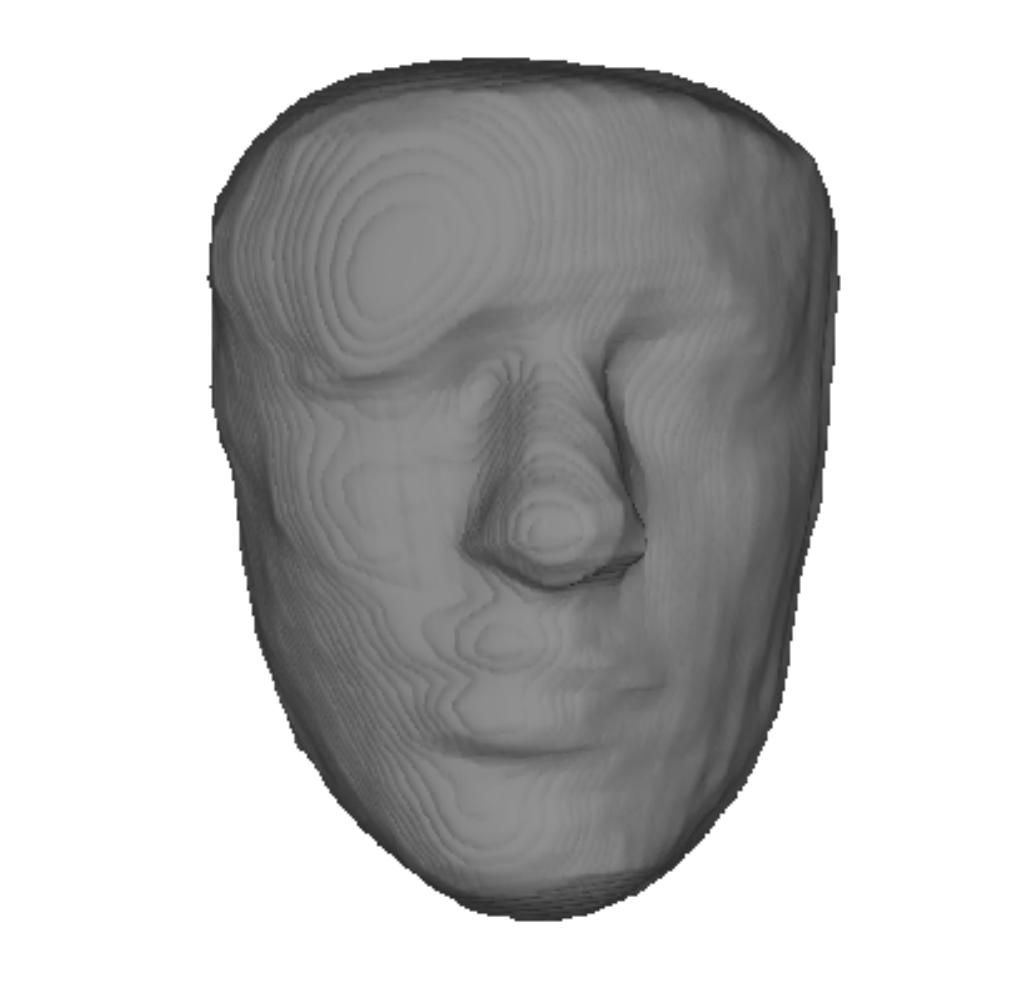}&
	\includegraphics[width=0.1538\textwidth]{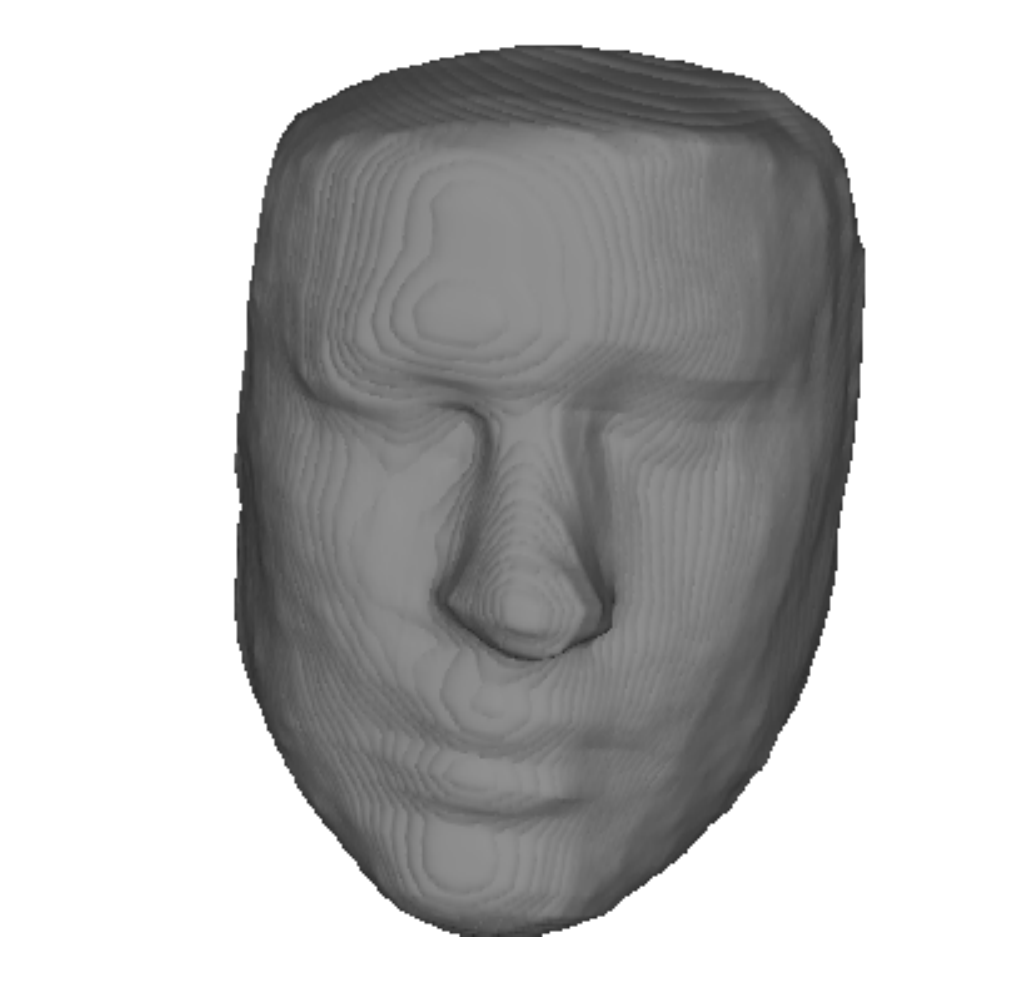}&
	\includegraphics[width=0.1538\textwidth]{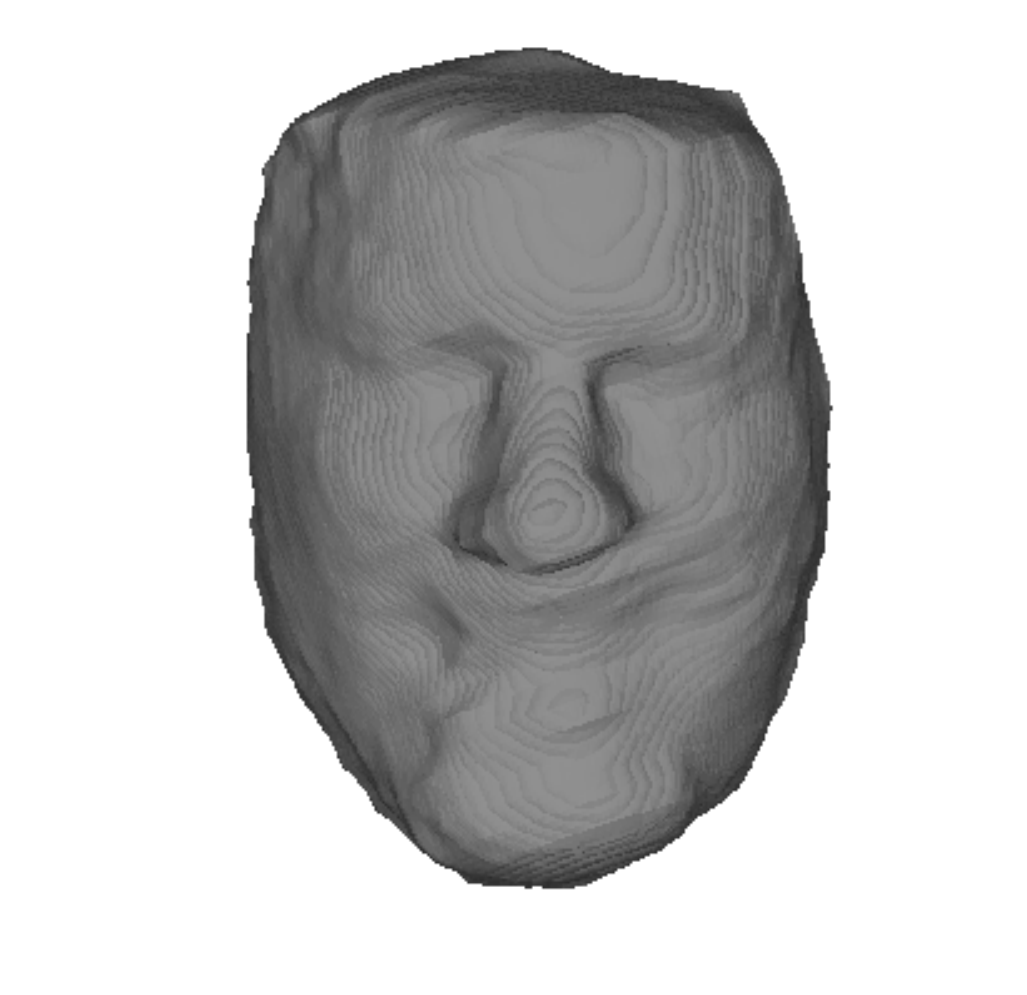}\\
    \centering textured model &
	\includegraphics[width=0.1538\textwidth]{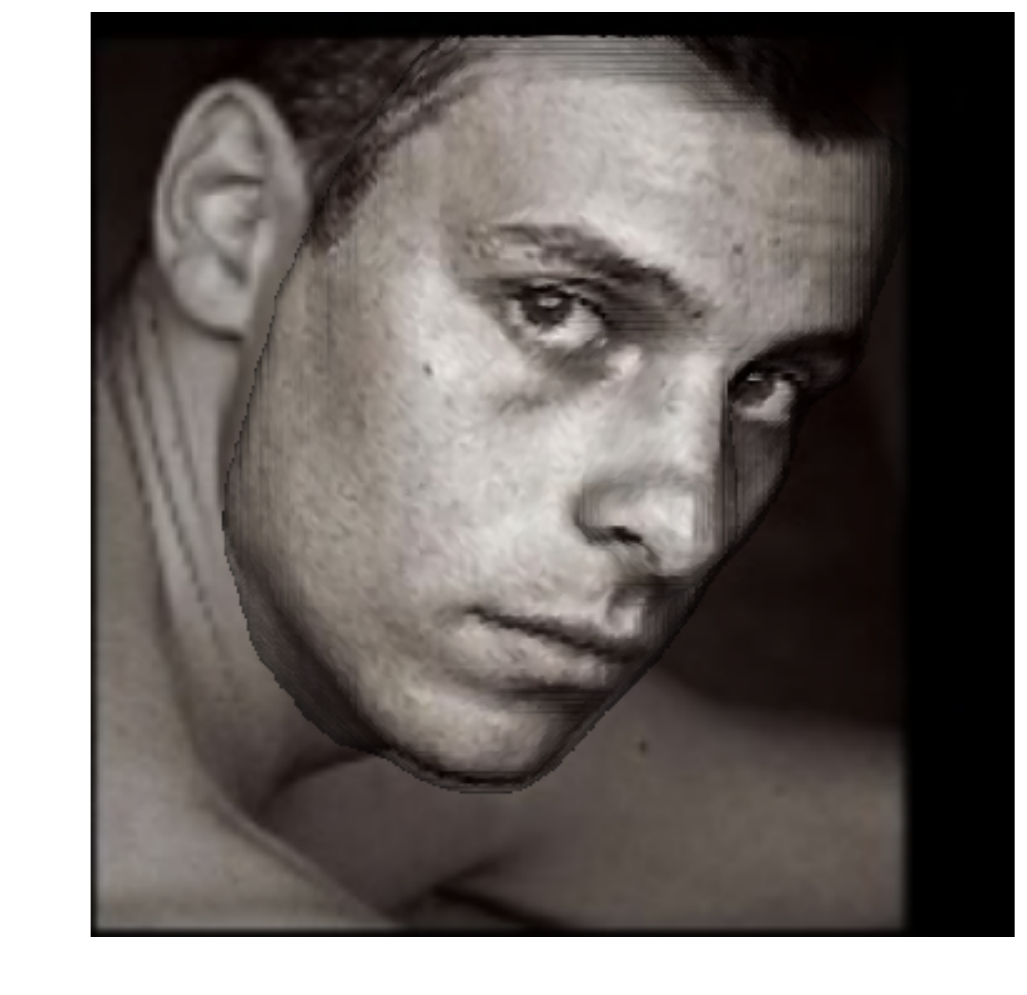}&
	\includegraphics[width=0.1538\textwidth]{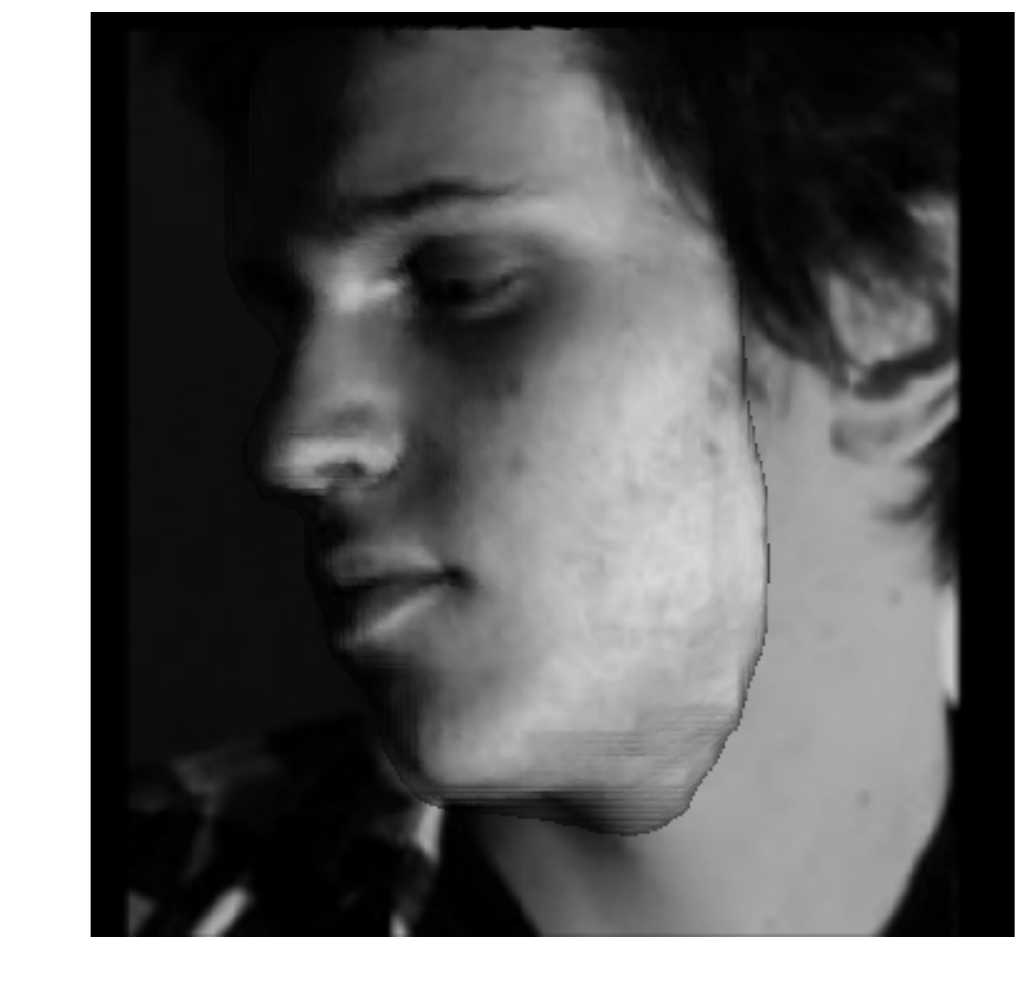}&
	\includegraphics[width=0.1538\textwidth]{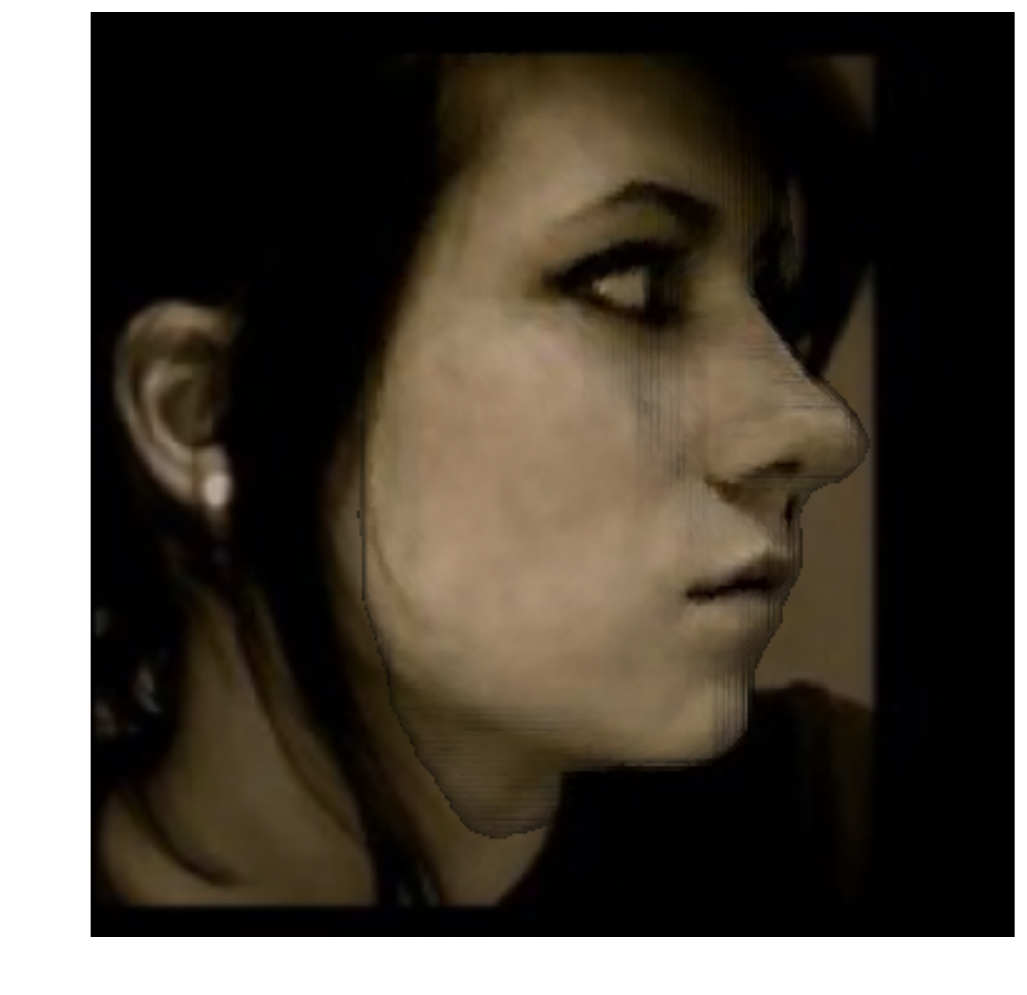}&
	\includegraphics[width=0.1538\textwidth]{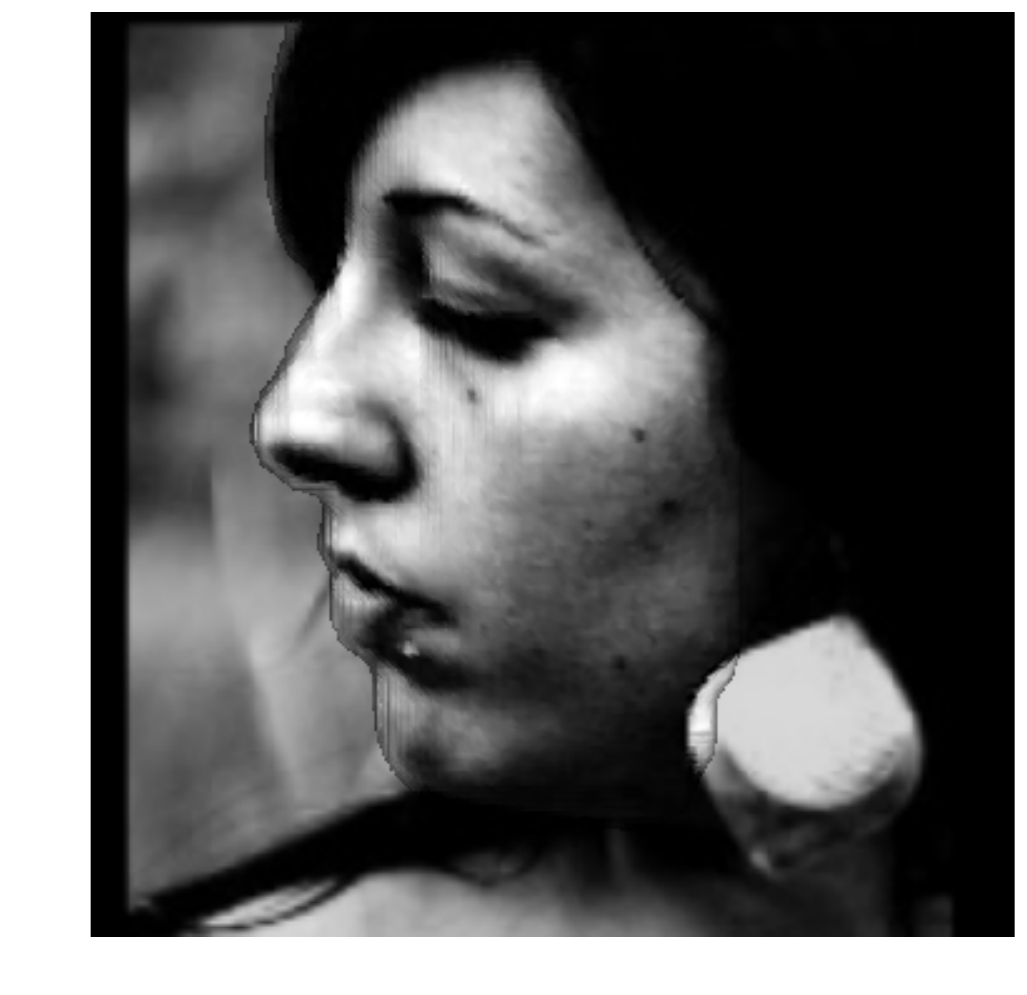}&
	\includegraphics[width=0.1538\textwidth]{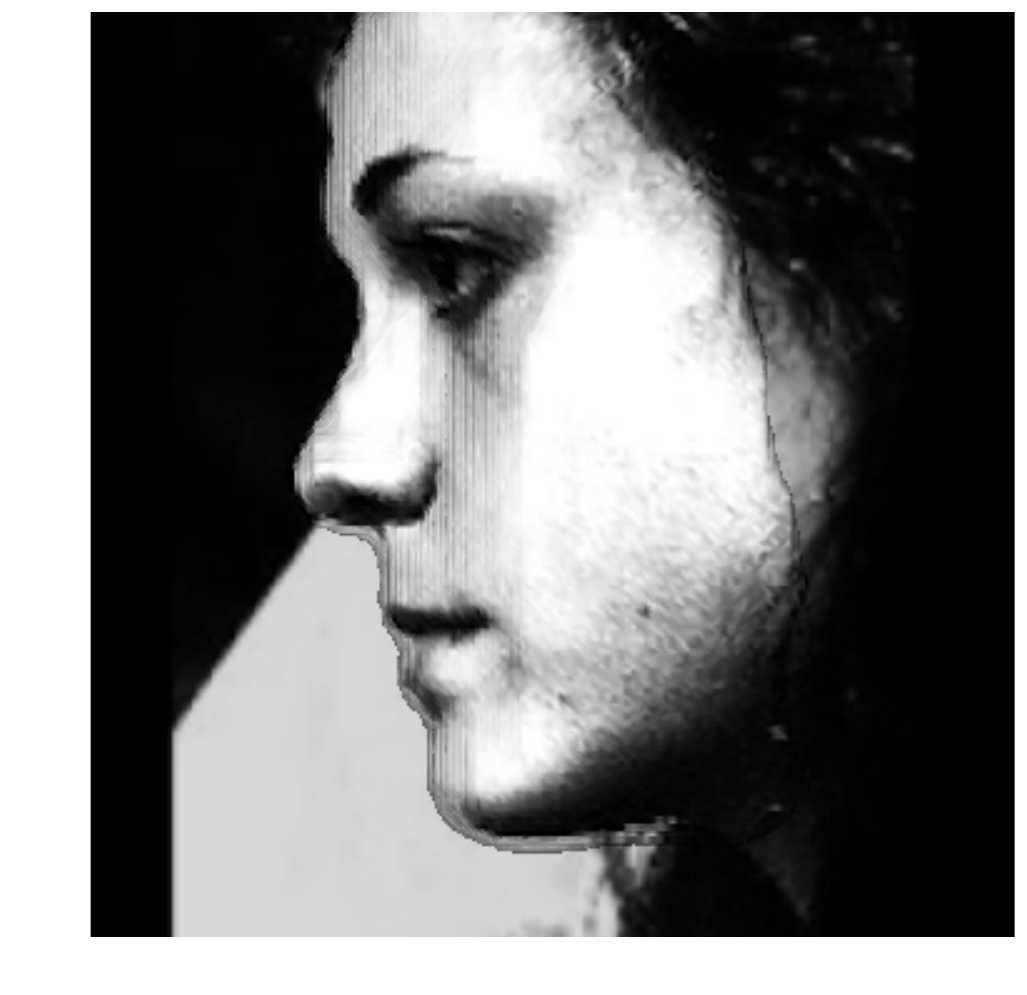}&
	\includegraphics[width=0.1538\textwidth]{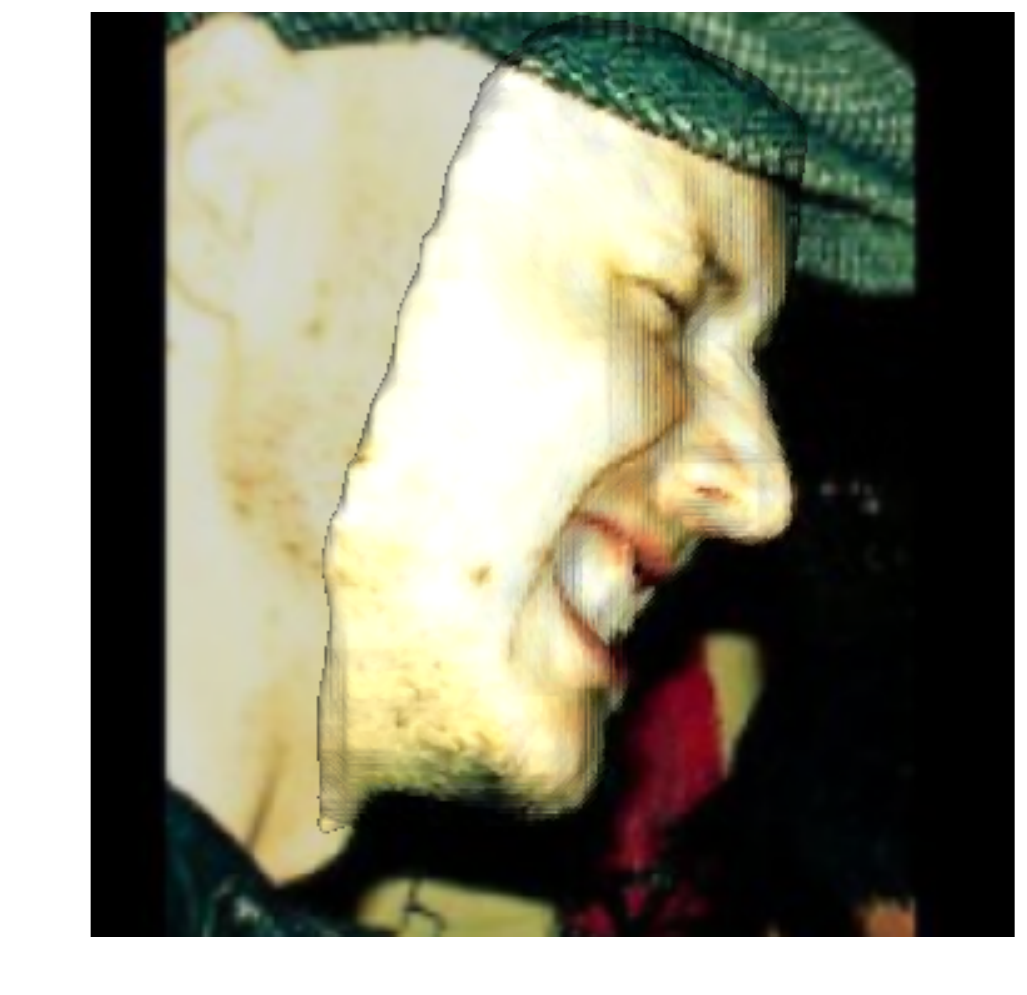}
  	\end{tabu}
    \caption{\label{fig:qualitative_AFLW2000} Qualitative results on AFLW2000 3D dataset. First row shows the input images; rows two and three show 3D models reconstructed using VRN without and with self-supervised bootstrap, rotated to a frontal viewpoint; fourth row shows the face images overlaid with texture-mapped reconstructed shapes using VRN with bootstrap.}
\end{figure*}

The AFLW2000-3D dataset \cite{DBLP:journals/corr/ZhuLLSL15} contains pairs of face images and 3D meshes for the first 2000 examples from AFLW \cite{koestinger11a}.
ICP was used to align the predicted and the ground-truth meshes, and the errors reported are after alignment. Reconstruction error in NME before and after applying the proposed self-supervised bootstrap method are shown in Fig. \ref{fig:nme_percent_AFLW2000} and Table \ref{tbl:quantitative_performance}.
Quantitative performance comparison in yaw orientation of the input face images is shown in the left column of Fig. \ref{fig:nme_yaw_pitch_roll}.
Qualitative results are shown in Fig.~\ref{fig:qualitative_AFLW2000}.
Fine-tuning VRN with bootstrapped data for only $\pm 20^{\circ}$ in yaw resulted in a small degradation in performance for AFLW2000-3D from 1.98\% to 2.02\%, but it has to be noted that AFLW2000-3D is biased towards frontal views.
Therefore, even large improvements at profile views can be overwhelmed by a small degradation at frontal views. In addition, a potential solution to mitigate the small degradation at the input image domain of frontal face images, which are not available to the fine-tuned network during bootstrap, could be the technique of learning without forgetting from \cite{DBLP:journals/corr/LiH16e}.
Bootstrapping with additional views with $\pm20^{\circ}$ and $\pm40^{\circ}$, and with $\pm20^{\circ}$, $\pm40^{\circ}$, and $\pm60^{\circ}$ in yaw angle decreases the average NME to 1.93\% and 1.90\%, respectively.

\subsection{Evaluation on the MICC Dataset\label{exp:300W}}
\begin{figure*}[bh]
    \begin{subfigure}{.5\textwidth}
    \centering
		\includegraphics[width=\linewidth]{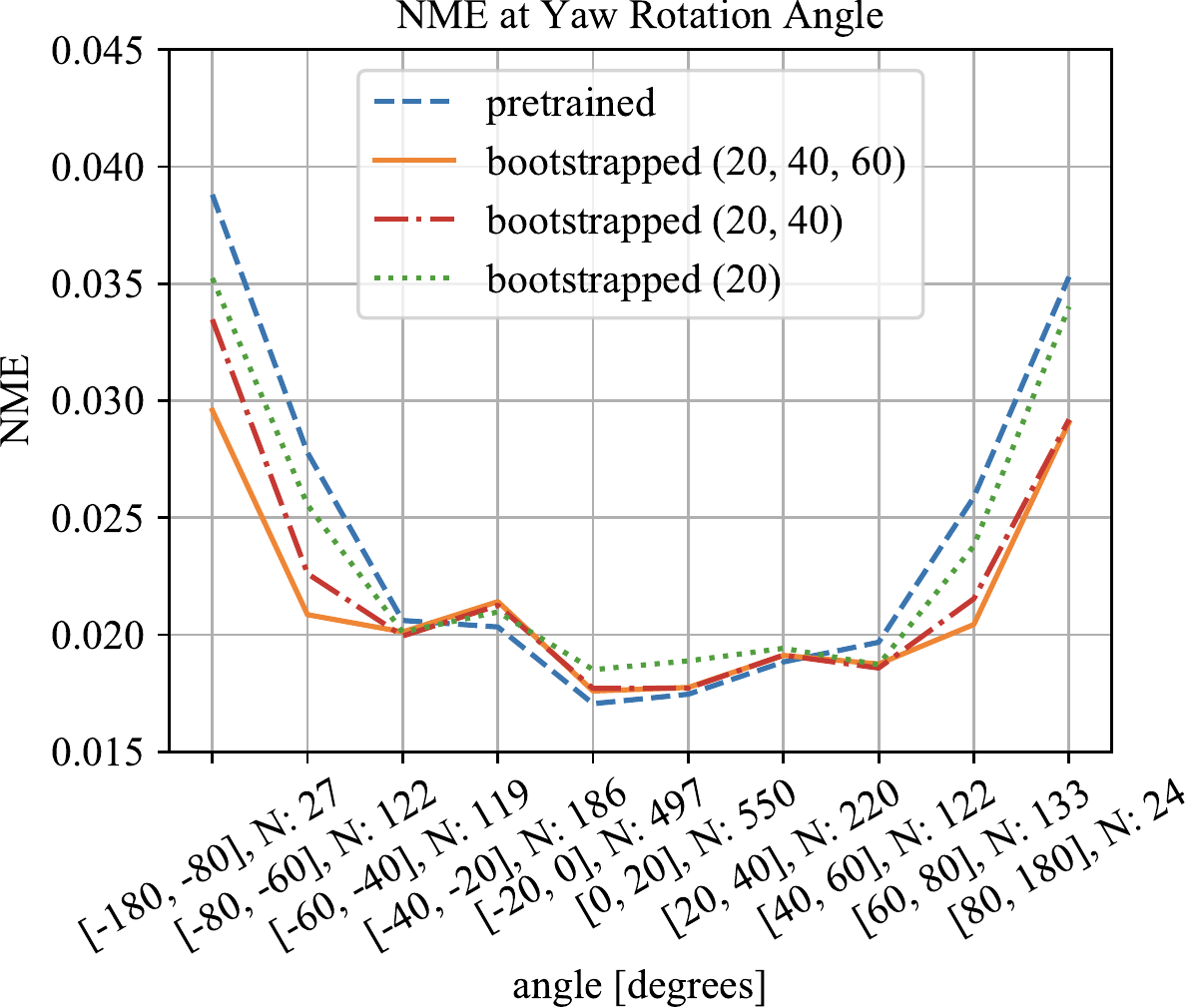} 
	\end{subfigure}%
	\hfill
	\begin{subfigure}{.5\textwidth}
    \centering
        \includegraphics[width=\linewidth]{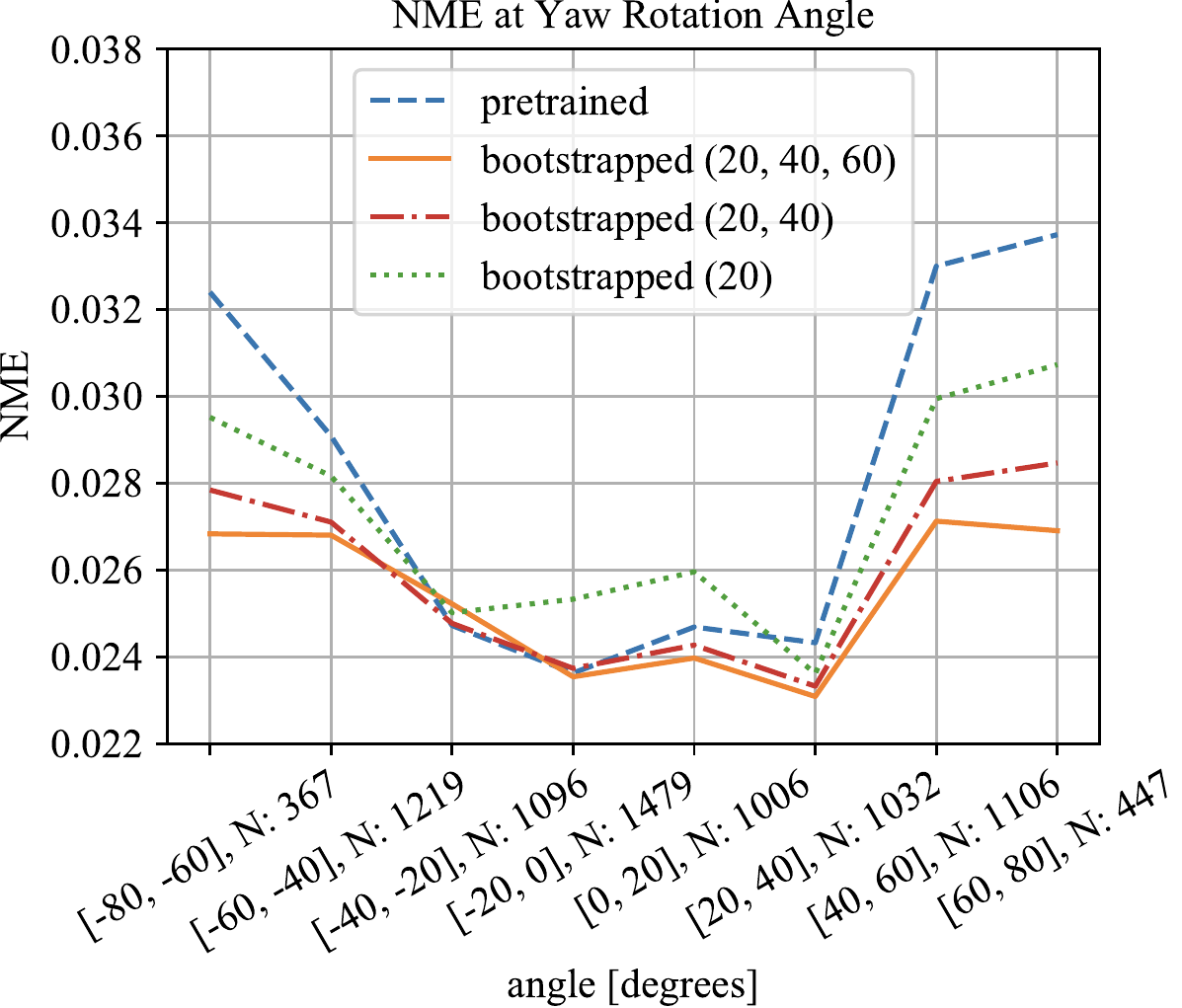} 
	\end{subfigure}%
	
    \caption{\label{fig:nme_yaw_pitch_roll} NME of the VRN-Unguided model with and without bootstrap in yaw angles of input images. Left column shows NME for the AFLW2000-3D dataset; right colum shows the error for the MICC (Cooperative) dataset.}
\end{figure*}

To fully demonstrate the effectiveness of the proposed method across large pose variations, it is desirable to evaluate it on a dataset with a more uniform distribution over facial poses.
Thus, we also report the reconstruction accuracy before and after bootstrapping on the MICC \cite{Bagdanov:2011:FHF:2072572.2072597} dataset, which contains a broader range of poses, as shown in Fig. \ref{fig:yaw_dist}.
The MICC dataset contains ground-truth 3D shapes for the faces of 53 subjects, acquired using structured-light scans.
It also contains 53 video sequences of the same subjects under varying resolutions and different types of environment.
We used the 53 Cooperative\footnote{Reconstruction accuracy of images in Indoor and Outdoor videos and more qualitative results on MICC are shown in the supplementary material.} videos (at the highest zoom level) and extracted in total 7752 frames for reconstruction. 
As with the previous experiment, ICP was used to align ground truth models and corresponding predicted shapes.  
Reconstruction accuracy in NME before and after bootstrapping with the application of different sets of rigid-body rotations are shown in Fig. \ref{fig:nme_percent_MICC} and Table. \ref{tbl:quantitative_performance}.
Steady improvement correlating with the addition of bootstrapped data is clear.
Furthermore, Fig.~\ref{fig:nme_yaw_pitch_roll} demonstrates that reconstruction error is significantly reduced at profile and near-profile views through the use of the proposed bootstrap method.

\section{Conclusions and Future Work}

We have developed a self-supervision method to improve the performance of deep models for single-image 3D face reconstruction.
Starting from a seeded set of data for which the network is known to have good performance, the proposed method generates new input-output pairs outside that original set by the controlled application of transformations in the 3D domain and rendering of the transformed 3D models.
The original network is then fine-tuned with data thus produced, and experimental results indicate that the method indeed improves the performance of the original model, particularly for inputs depicting challenging viewpoints with large yaw or pitch angles.

As future work, we will investigate the impact of the proposed method on the fine-tuning of face-reconstruction models other than VRN, including 3DMM methods and extend the transformation $T$ beyond the Euclidean group of rigid motions, to a larger class of transformations including changes in illumination and facial expressions.
We will also consider the application of the method to other 3D reconstruction problems, beyond faces.
{\small
\printbibliography}

\end{document}